%% file: templateArxiv.tex
\documentclass{article}
\usepackage{amsmath}   % 支持数学公式和向量符号
\usepackage{amssymb}   % 支持 \mathbb{R} 等特殊数学符号
\usepackage{microtype}  % 改善文字间距和排版，减少underfull/overfull警告
\usepackage{colortbl}   % 单元格颜色控制
\usepackage{subcaption} % for subfigure environment
\usepackage{xcolor}     % 颜色定义
\usepackage{booktabs}   % 表格线美化（原表依赖）
\usepackage{graphicx}   % 缩放功能（原表依赖）
\usepackage{nicefrac}       % compact symbols for 1/2, etc.
\usepackage{amsmath}
\usepackage{PRIMEarxiv}
\usepackage{lipsum}
\usepackage{subcaption} % for subfigure environment
\usepackage[utf8]{inputenc} % allow utf-8 input
\usepackage[T1]{fontenc}    % use 8-bit T1 fonts
\usepackage{hyperref}       % hyperlinks
\usepackage{url}            % simple URL typesetting
\usepackage{booktabs}       % professional-quality tables
\usepackage{amsfonts}       % blackboard math symbols
\usepackage{nicefrac}       % compact symbols for 1/2, etc.
\usepackage{microtype}      % microtypography
\usepackage{lipsum}
\usepackage{fancyhdr}       % header
\usepackage{graphicx}       % graphics
\graphicspath{{media/}}     % organize your images and other figures under media/ folder

\title{SYN-DIAG: An LLM-based Synergistic Framework for Generalizable Few-shot Fault Diagnosis on the Edge
}

\author{
  Zijun Jia \\
  Beihang University \\
  \texttt{21375166@buaa.edu.cn} \\
  \And
  Shuang Liang \\
  Hangzhou International Innovation Institute of Beihang University \\
  \texttt{liangshuangls@buaa.edu.cn} \\
  \And
  Jinsong Yu \\
 Beihang University \\
  \texttt{yujs@buaa.edu.cn} \\
}

\begin{document}
\maketitle

\begin{abstract}
Industrial fault diagnosis faces the dual challenges of data scarcity and the difficulty of deploying large AI models in resource-constrained environments. This paper introduces Syn-Diag, a novel cloud-edge synergistic framework that leverages Large Language Models to overcome these limitations in few-shot fault diagnosis. Syn-Diag is built on a three-tiered mechanism: 1) Visual-Semantic Synergy, which aligns signal features with the LLM's semantic space through cross-modal pre-training; 2) Content-Aware Reasoning, which dynamically constructs contextual prompts to enhance diagnostic accuracy with limited samples; and 3) Cloud-Edge Synergy, which uses knowledge distillation to create a lightweight, efficient edge model capable of online updates via a shared decision space. Extensive experiments on six datasets covering different CWRU and SEU working conditions show that Syn-Diag significantly outperforms existing methods, especially in 1-shot and cross-condition scenarios. The edge model achieves performance comparable to the cloud version while reducing model size by 83\% and latency by 50\%, offering a practical, robust, and deployable paradigm for modern intelligent diagnostics.
\end{abstract}

% keywords can be removed
\input{Introduction}

\input{releated_works}
\input{method}

\input{Experiment_seting}
\input{Experiment}
\input{Conclusion}

% Bibliography
\bibliographystyle{unsrt}  
\bibliography{references}

\end{document}

%% file: Introduction.tex
\section{Introduction}

In the wave of the Fourth Industrial Revolution (Industry 4.0), advanced manufacturing, centered on intelligence and automation, is becoming a key engine for global economic growth. Among its core components, the health status monitoring and fault diagnosis of mechanical equipment serve as the cornerstone for ensuring the continuous, efficient, and safe operation of production lines~\cite{zhu2023review}. Any sudden equipment failure can interrupt complex production processes, leading not only to significant economic losses but also to potentially catastrophic safety accidents. Therefore, developing intelligent diagnostic systems capable of early, accurate, and automatic identification of equipment faults is of crucial theoretical and practical significance for enhancing production efficiency, reducing maintenance costs, and ensuring system reliability~\cite{zhou2022towards}.
\begin{figure}
    \centering
    \includegraphics[width=0.8\linewidth]{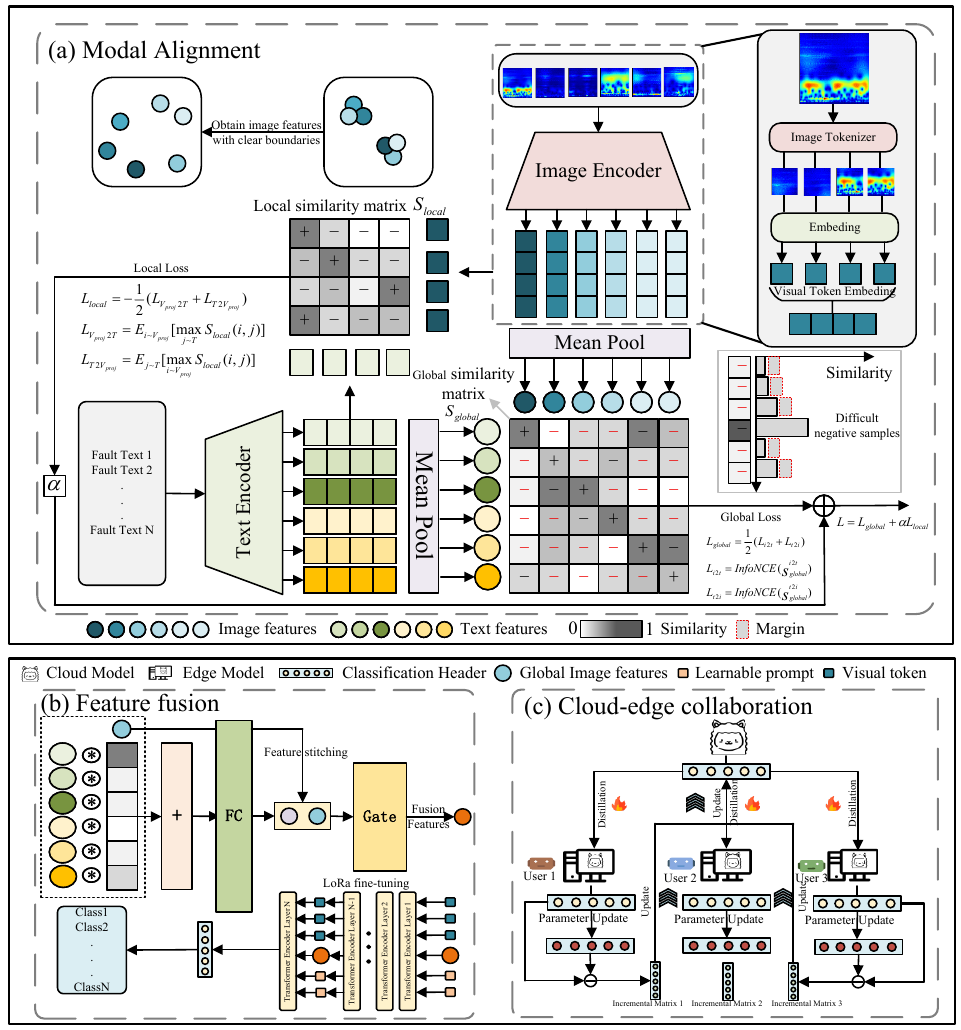}
    \caption{Overall architecture diagram}
    \label{fig:placeholder}
\end{figure}
In recent years, artificial intelligence technologies, represented by Deep Learning, have made remarkable progress in the field of fault diagnosis. From classic Convolutional Neural Networks (CNNs)~\cite{qin2022fault} and Deep Residual Networks (ResNets)~\cite{yin2024bi} to the recently outstanding Vision Transformers (ViTs) and their variants~\cite{ji2024devit}, these models achieve end-to-end fault classification by automatically learning features from vibration signals, acoustic signals, or their transformed time-frequency spectrograms. However, these purely data-driven methods generally face two core challenges in practical industrial applications. First, data scarcity: the performance of deep learning models is highly dependent on large-scale, diverse, and well-labeled training data. In industrial settings, however, acquiring fault samples (especially for rare or severe faults) is both difficult and expensive, often leading to model overfitting and poor generalization due to insufficient training data~\cite{jiang2025adaptive}. Second, deployment difficulty: in the pursuit of high accuracy, model architectures have become increasingly large and complex. Parameter counts in the billions make their computational resource requirements (such as GPU memory and processing power) extremely high, rendering the direct deployment of these large models on resource-constrained edge computing devices impractical~\cite{pan2024cloud}.

To address these challenges, academia and industry are actively exploring new technological paradigms. The rise of Large Language Models (LLMs) has brought new hope for resolving this dilemma~\cite{he2025full,zhao2025adjust,ran2025fd}. Through their massive pre-training data, LLMs have not only acquired extensive linguistic knowledge but have also internalized powerful logical reasoning and few-shot learning capabilities. Recently, applying LLMs to the field of fault diagnosis has emerged as a new and promising research direction. Researchers have attempted to use LLMs to process numerical sensor data~\cite{lin2025fd,zheng2024empirical,xiao2025krail} or to construct LLM-based multi-model collaborative frameworks to solve link prediction problems in evolving fault graphs~\cite{wang2025multi}. In particular, the further development of Vision-Language Models (VLMs), such as CLIP~\cite{radford2021learning}, has successfully combined visual perception with powerful semantic understanding through contrastive learning on massive image-text pairs. This progress has inspired researchers to apply multimodal reasoning to fault diagnosis, for instance, by fine-tuning models with visual instruction data containing signal images and expert knowledge to achieve explainable diagnostics~\cite{wang2025diagllm}. We believe that this ability to integrate visual signals with human expert knowledge (in textual form) offers a novel approach to overcoming the challenge of data scarcity.

Inspired by this, this paper proposes a novel cloud-edge synergistic fault diagnosis framework named "Synergistic Intelligence Diagnosis" (Syn-Diag), which aims to systematically address the dual challenges of data scarcity and deployment difficulty. The core advantage of this framework lies in its unique mechanism for handling few-shot problems. Through a cross-modal pre-training stage of vision-semantic synergy, we deeply align the visual features of time-frequency spectrograms with the textual descriptions of fault mechanisms embedded in the LLM. In this stage, we design an innovative multi-level alignment loss function that quadratically enriches the supervisory signals by constructing dense positive and negative sample pairs at both global and local levels. This forces the model to learn highly discriminative feature representations from an extremely small amount of labeled data. Subsequently, we design a synergistic content-aware inference mechanism. This mechanism first dynamically compares the input image with the textual descriptions of all candidate faults, adaptively generating a context-aware hybrid representation through a Gated Fusion Unit. Then, we introduce Deep Learnable Prompts to guide the LLM at multiple layers, structuring the inference process like a multiple-choice question, which significantly enhances the model's discriminative ability under few-shot conditions. Finally, through a cloud-edge synergistic framework of knowledge distillation and online updating, we efficiently transfer the knowledge from the large cloud model (Syn-Diag-Cloud) to a lightweight edge model (Syn-Diag-Edge) and enable it to perform low-cost online updates to the cloud model using new data from the edge.

To validate the effectiveness of our framework, we conducted comprehensive experiments on two widely used public datasets: the Case Western Reserve University (CWRU) bearing dataset and the Southeast University (SEU) gearbox dataset. The experimental results show that under strict few-shot (1-shot, 3-shot, 5-shot, 7-shot) and Cross-Condition settings, our proposed model significantly outperforms existing baseline methods across multiple metrics, including accuracy, precision, and F1-score. Notably, even in the extreme case of only a single training sample, our model maintains near-perfect diagnostic performance. Furthermore, our edge model achieves performance highly consistent with the cloud model, yet its model size, GPU memory footprint, and inference latency are reduced by several orders of magnitude, fully demonstrating its feasibility for practical deployment.

The main contributions of this paper can be summarized as follows: 1) We propose an innovative, LLM-based synergistic intelligence framework for fault diagnosis that effectively fuses visual information and semantic knowledge, significantly improving few-shot diagnostic performance. 2) We design and implement a complete technical solution that includes multi-level alignment, synergistic inference, and knowledge distillation, systematically addressing the entire chain of challenges from model training to edge deployment. 3) We propose and validate an efficient cloud-edge synergistic online updating mechanism, providing a new paradigm for building industrial intelligent systems capable of continuous learning and adaptive evolution. The remainder of this paper is organized as follows: Chapter 2 reviews related work, Chapter 3 details our proposed method, Chapter 4 describes the experimental setup, Chapter 5 presents and analyzes the experimental results, and Chapter 6 concludes the paper and discusses future work.

%% file: releated_works.tex
\section{Related Works}
\subsection{Few-shot Fault Diagnosis}

Few-shot Fault Diagnosis (FSFD) is a critical research topic in the field of industrial intelligence. Its core objective is to build high-performance diagnostic models when only a very small number of labeled samples are available for each fault category. This challenge stems directly from industrial realities, where fault data, especially for rare and critical failures, is difficult to obtain. To address this data scarcity issue, researchers have primarily explored three technical paths.

The first path is based on Data Augmentation. These methods aim to expand the limited training set through transformation or generation techniques. Traditional methods include rotating signals, adding noise, or applying time-domain transformations~\cite{zhao2020deep}. In recent years, with the development of Generative Adversarial Networks (GANs), researchers have begun to use GANs and their variants to generate high-quality, diverse synthetic fault signals, thereby alleviating the problem of data insufficiency~\cite{li2022multi, sifat2025gan}. Although data augmentation has improved model performance to some extent, the diversity and fidelity of the generated samples remain a technical bottleneck, and it is difficult to fully simulate the complexity of real-world faults.

The second path is based on Transfer Learning. The core idea is to transfer knowledge learned from a data-rich source domain (e.g., laboratory simulation data) to a data-scarce target domain (e.g., actual operational condition data)~\cite{an2023gaussian,xia2025fault,zhang2025adaptive}. Domain Adaptation is a mainstream technique within this path, which minimizes the distribution discrepancy between the source and target domains in the feature space, enabling the model to apply the discriminative ability learned in the source domain to the target domain~\cite{zhu2020new}. However, the success of transfer learning is highly dependent on the correlation between the source and target domains. When the difference between them is too large (i.e., a Domain Shift occurs), performance can drop sharply~\cite{qian2024adaptive}.

The third path is based on Meta-Learning, also known as "learning to learn." Meta-learning aims to learn a general learning strategy or a good initial model state from multiple related diagnostic tasks, enabling it to quickly adapt to new diagnostic tasks using only a few samples~\cite{che2022few,liang2025interpretable}. Among these, methods based on Metric Learning, such as Prototypical Networks and Siamese Networks, learn a metric space where samples of the same class are closer and samples of different classes are farther apart, thereby achieving few-shot classification~\cite{zhang2022supervised, wu2020few}. Although meta-learning has achieved significant success in few-shot classification, most of these methods are still limited to learning from visual or signal features themselves. They fail to effectively leverage external knowledge and have limited capabilities when faced with faults that are semantically complex and difficult to distinguish based on visual features alone.

\subsection{Multimodal Large Language Models}

Large Language Models, such as the GPT series~\cite{brown2020language}, have demonstrated powerful capabilities in natural language understanding, generation, and reasoning through pre-training on massive text corpora, leading to paradigm shifts across various industries. Recently, this trend has extended to the multimodal domain, giving rise to Multimodal Large Language Models (MLLMs), which can simultaneously process and correlate information from different modalities, such as text, images, and audio.

A cornerstone of MLLMs is Vision-Language Pre-training (VLP). Among VLP techniques, contrastive learning methods, represented by CLIP~\cite{radford2021learning}, learn to map images and their corresponding text descriptions into a unified embedding space by training on a dataset containing hundreds of millions of image-text pairs. This approach enables the model to perform image classification tasks in a "zero-shot" manner simply by providing textual descriptions of the categories, showcasing unprecedented generalization capabilities. Subsequent works such as LLaVA~\cite{liu2023visual} and InstructBLIP~\cite{dai2023instructblip} have further enhanced the ability of these models to follow complex natural language instructions and perform diverse vision-language tasks through Instruction Tuning. To efficiently adapt these massive models for downstream tasks, Parameter-Efficient Fine-Tuning (PEFT) techniques like Low-Rank Adaptation (LoRA) are widely adopted~\cite{hu2022lora}.

Applying the powerful capabilities of LLMs and MLLMs to the field of industrial fault diagnosis has become a promising and emerging research direction, attracting significant attention. Some pioneering works have begun to explore this possibility, mainly following two technical routes. The first route focuses on leveraging the outstanding knowledge integration and logical reasoning abilities of LLMs. For instance, Liu et al. explored combining knowledge graphs with an LLM for fault diagnosis in the aerospace assembly domain, enhancing diagnostic accuracy and interpretability by fusing structured knowledge with the LLM's reasoning~\cite{peifeng2024joint}. Other studies have utilized LLMs for test-free fault localization, demonstrating their potential in understanding and reasoning about complex system logic~\cite{yang2024large}. The second route is dedicated to enabling LLMs to directly process and understand non-textual data from industrial sites; for example, Tao et al. designed a specialized framework to apply an LLM directly to bearing fault diagnosis~\cite{tao2025llm}. More relevant to our work is the trend of combining these two capabilities—that is, leveraging multimodal capabilities to fuse signal data (visual modality) with expert knowledge (text modality). Together, these studies reveal the great potential of using the semantic understanding and reasoning capabilities of LLMs to enhance fault diagnosis models, especially in addressing data scarcity and improving model interpretability.

However, existing work has largely focused on the direct application of model capabilities, while research that systematically addresses the entire chain of problems from pre-training and few-shot fine-tuning to final edge deployment is still lacking. The "Synergistic Intelligence Diagnosis" framework proposed in this paper is situated in this context, aiming to build an end-to-end solution. We not only draw upon the ideas of contrastive learning and instruction tuning but also design a series of innovative mechanisms—such as a multi-level alignment loss, synergistic content-aware reasoning, and cloud-edge collaborative knowledge distillation with online updates. Through these innovations, we aim to systematically address the core challenges of few-shot learning and practical deployment in industrial intelligence while ensuring state-of-the-art (SOTA) performance.

%% file: method.tex
\section{Method}

\subsection{Cross-modal Semantic Alignment Pre-training}
\label{subsec:method_alignment}

To enable the LLM to comprehend visual information in fault diagnosis tasks, we have designed a critical cross-modal semantic alignment pre-training stage. As shown in Figure~\ref{fig:placeholder}~(a), the core objective of this stage is to construct a mapping from the visual feature space to the LLM's native semantic space, allowing the model to associate input time-frequency spectrograms with their corresponding fault description texts. To preserve the strong prior knowledge of LLM and improve training efficiency, we freeze its parameters and only fine-tune a lightweight visual feature extractor for efficient knowledge alignment.

\subsubsection{Model Architecture}
Our alignment model primarily consists of a Visual Feature Extractor and a pre-trained LLM. The purpose of the visual feature extractor $f_v(\cdot)$ is to convert an input 2D time-frequency spectrogram $I \in \mathbb{R}^{H \times W \times C}$ into a series of visual feature embeddings compatible with the LLM's hidden layer dimensions. We employ an architecture based on a CNN, which transforms a $224 \times 224 \times 3$ image into a $7 \times 7$ feature map with 256 channels. To align with the LLM's sequential processing style, we reshape this feature map into a visual feature sequence $V \in \mathbb{R}^{N_v \times D_{\text{cnn}}}$ (sequence length $N_v=49$) and map its dimension to the LLM's hidden dimension $D_{\text{llm}}$ using a linear projection layer $W_p$:
\begin{equation}
    V_{\text{proj}} = f_v(I) = V \cdot W_p, \quad \text{where } V_{\text{proj}} \in \mathbb{R}^{N_v \times D_{\text{llm}}}.
\end{equation}
On the semantic side, we generate descriptive text prompts for each fault category and convert them into ID sequences using the LLM's tokenizer. The semantic features $T \in \mathbb{R}^{N_t \times D_{\text{llm}}}$ are obtained directly from the LLM's word embedding layer $f_t(\cdot)$. Since the word embedding layer is frozen during the pre-training phase, it provides a stable and high-quality semantic space that serves as an "anchor point" for visual feature alignment.

\subsubsection{Multi-level Alignment Loss Function}
To achieve robust and fine-grained cross-modal alignment, we designed a hybrid loss function that combines both global and local levels. This design ensures that the model not only understands the correspondence between images and texts as a whole but also captures their association at a fine-grained feature level.

\textbf{Global Contrastive Loss with Hard Negative Mining.} At the global level, we employ the InfoNCE contrastive learning loss with Hard Negative Mining. First, we obtain global feature vectors $\mathbf{v}_g$ and $\mathbf{t}_g$ by applying average pooling to the visual and semantic feature sequences, followed by L2 normalization. For a batch of size $B$, the image-to-text contrastive loss $\mathcal{L}_{\text{i2t}}$ is defined as:
\begin{equation}
\label{eq:2}
    \mathcal{L}_{\text{i2t}} = -\frac{1}{B} \sum_{i=1}^{B} \log \frac{\exp(\text{sim}(\mathbf{v}_{g,i}, \mathbf{t}_{g,i}) / \tau)}{\sum_{j=1}^{B} \exp(\text{sim}(\mathbf{v}_{g,i}, \mathbf{t}_{g,j}) / \tau)},
\end{equation}
where $\tau$ is a learnable temperature coefficient. The text-to-image loss $\mathcal{L}_{\text{t2i}}$ is defined symmetrically. The global loss is $\mathcal{L}_{\text{global}} = (\mathcal{L}_{\text{i2t}} + \mathcal{L}_{\text{t2i}}) / 2$. To focus the model on distinguishing confusable samples, we incorporate a hard negative mining strategy within the contrastive loss. For each anchor sample (e.g., an image's global feature $\mathbf{v}_{g,i}$), we first identify the hardest negative sample from the batch, which is the text sample $\mathbf{t}_{g,j}$ ($j \neq i$) with the highest similarity score, denoted as $s^{*}_{i} = \max_{j \neq i} \text{sim}(\mathbf{v}_{g,i}, \mathbf{t}_{g,j})$. We then use this score to dynamically suppress the influence of easy negatives by penalizing any negative sample whose similarity score is significantly lower than this hardest negative. This is achieved by defining a modified similarity score, $s'_{ik}$, for all negative pairs ($k \neq i$):
\begin{equation}
    s'_{ik} = \text{sim}(\mathbf{v}_{g,i}, \mathbf{t}_{g,k}) - m_{\text{dyn}} \cdot \mathbb{I}\left(\text{sim}(\mathbf{v}_{g,i}, \mathbf{t}_{g,k}) < s^{*}_{i} - m_{\text{dyn}}\right),
\end{equation}
where $\mathbb{I}(\cdot)$ is the indicator function. The effectiveness of this penalty lies in its interaction with the loss calculation in Equation~\ref{eq:2}. By subtracting the margin $m_{\text{dyn}}$ from the logits of simple negative samples, their contribution to the softmax denominator decreases exponentially, becoming negligible. This mechanism effectively reshapes the gradient landscape, forcing the optimization to concentrate on the narrow, more challenging decision boundary between the positive pair and the hard negatives. The margin $m_{\text{dyn}}$ itself is not fixed but smoothly increases from 0.1 to 0.7 using a cosine scheduler as training progresses, compelling the model to tackle progressively more difficult distinctions in later stages.

\textbf{Local Maximum Similarity Loss.} Global alignment alone may overlook the association between local regions of an image and specific words in the text. To address this, we propose a Local Maximum Similarity Loss ($\mathcal{L}_{\text{local}}$), which encourages each token in the visual sequence to find a strongly corresponding token in the text sequence, and vice versa. We first compute the cosine similarity for all token pairs between the visual sequence $V_{\text{proj}}$ and the text sequence $T$, yielding a local similarity matrix $S_{\text{local}} \in \mathbb{R}^{N_v \times N_t}$. The loss is defined as the negative mean of the bidirectional maximum similarities:
\begin{equation}
\label{eq:local_loss_condensed}
\mathcal{L}_{\text{local}} = -\frac{1}{2} \left( \frac{1}{N_v} \sum_{i=1}^{N_v} \max_{j} S_{\text{local}}(i, j) + \frac{1}{N_t} \sum_{j=1}^{N_t} \max_{i} S_{\text{local}}(i, j) \right).
\end{equation}
Minimizing $\mathcal{L}_{\text{local}}$ incentivizes the model to learn a more refined, token-level alignment, which is crucial for understanding fault details such as location, type, and severity.

\textbf{Overall Objective Function.} Finally, our total alignment loss function $\mathcal{L}_{\text{align}}$ is a weighted sum of the global contrastive loss and the local maximum similarity loss:
\begin{equation}
    \mathcal{L}_{\text{align}} = \mathcal{L}_{\text{global}} + \alpha \mathcal{L}_{\text{local}},
\end{equation}
where $\alpha$ is a hyperparameter that balances the importance of global and local alignment.

\subsubsection{Advantages and Feature Generalization}
Our proposed multi-level alignment method offers significant advantages. First, the contrastive learning framework greatly improves sample utilization efficiency. In traditional classification tasks, each sample is only associated with its own label. In our framework, by constructing positive pairs (matching image-text) and negative pairs (mismatched image-text), the model compares every sample with all other samples in the batch. A batch of size $B$ generates $B$ positive pairs and $B(B-1)$ negative pairs, achieving a quadratic-level exploration of inter-sample relationships. This dense supervisory signal compels the model to learn a structured and semantically distinguishable embedding space. Second, by combining hard negative mining with local detail alignment, the model is guided to focus on the most discriminative subtle features rather than merely memorizing superficial patterns. Through t-SNE visualization analysis, as shown in Figure~\ref{fig:CWRU_SNE},~\ref{fig:SEU_SNE}, we observe that the visual features after alignment training exhibit significant semantic clustering—fault images of the same category cluster tightly in the feature space, while clear boundaries form between different categories. This naturally emerging clustering structure demonstrates that our alignment method successfully injects semantic information into the visual features, making the feature representation both retain visual patterns and embody the semantic concepts of fault categories. Ultimately, this high-quality pre-training lays a solid foundation for subsequent few-shot fault diagnosis tasks, as the powerful cross-modal prior knowledge acquired by the model enables it to generalize rapidly and identify accurately with extremely high data efficiency when faced with new categories or data-scarce scenarios.

\subsection{Dynamic Synergistic Prompt Fusion and Deep Prompt Tuning}
\label{subsec:method_fusion_prompting}

After completing the cross-modal semantic alignment, the model enters the few-shot fine-tuning stage, as depicted in Figure~\ref{fig:placeholder}~(b). The core of this stage is to achieve dynamic, content-aware reasoning. To this end, we have designed a Dynamic Synergistic Prompt Fusion mechanism, combined with Deep Prompt Tuning, to maximally stimulate the LLM's contextual understanding and reasoning capabilities with minimal parameter overhead.

\subsubsection{Dynamic Synergistic Prompt Fusion}
Traditional methods typically match image features with a single text description, ignoring potential correlations between different fault categories. Our framework, in contrast, constructs a "multiple-choice" style context containing all possible options, forcing the model to make a decision through comparison. Specifically, for an input time-frequency spectrogram, we first extract its global visual feature $\mathbf{v}_g \in \mathbb{R}^{D_{\text{llm}}}$. Concurrently, we retrieve the text description embeddings for all $C$ candidate fault categories in the dataset, $\{\mathbf{T}_1, \mathbf{T}_2, \dots, \mathbf{T}_C\}$, and compute their corresponding global semantic features $\{\mathbf{t}_{g,1}, \mathbf{t}_{g,2}, \dots, \mathbf{t}_{g,C}\}$. The model calculates the cosine similarity between $\mathbf{v}_g$ and each category's semantic feature $\mathbf{t}_{g,c}$, converting these similarities into a set of normalized attention weights $\{w_1, w_2, \dots, w_C\}$ via a Softmax function:
\begin{equation}
    w_c = \frac{\exp(\text{sim}(\mathbf{v}_g, \mathbf{t}_{g,c}))}{\sum_{j=1}^{C} \exp(\text{sim}(\mathbf{v}_g, \mathbf{t}_{g,j}))}.
\end{equation}
These weights reflect the degree of relevance between the input image and each fault category. Subsequently, we use these weights to perform a weighted sum of all category text embedding sequences, generating a dynamic, content-aware mixed text embedding $\mathbf{T}_{\text{mix}} \in \mathbb{R}^{N_t \times D_{\text{llm}}}$:
\begin{equation}
    \mathbf{T}_{\text{mix}} = \sum_{c=1}^{C} w_c \mathbf{T}_{c}.
\end{equation}
Finally, we feed the global visual feature $\mathbf{v}_g$ and the global feature of the mixed text $\mathbf{t}_{g,\text{mix}}$ (obtained by average pooling $\mathbf{T}_{\text{mix}}$) into a Gated Fusion Unit (GFU). This unit adaptively fuses the information from both modalities using a learnable gating vector $\mathbf{z}$, producing a unified multimodal fused feature $\mathbf{h}_{\text{fused}}$, which serves as the starting point for subsequent processing by the LLM.

\subsubsection{Deep Prompt Tuning}
To efficiently guide the LLM to adapt to the fault diagnosis task in a few-shot scenario while avoiding catastrophic forgetting, we employ Deep Prompt Tuning. Unlike traditional methods that only add prompts at the input layer, this technique injects learnable prompts at multiple layers of the LLM. Specifically, we design two sets of prompts.

The first set is the Input Layer Prompts $\mathbf{P}_{\text{in}} \in \mathbb{R}^{L_{\text{in}} \times D_{\text{llm}}}$: a set of learnable vectors prepended to the multimodal fused feature $\mathbf{h}_{\text{fused}}$ and the visual feature sequence $V_{\text{proj}}$, which together form the initial input sequence for the LLM.

The second set consists of Inter-layer Prompts $\mathbf{P}_{l} \in \mathbb{R}^{L_{\text{deep}} \times D_{\text{llm}}}$: at specific pre-defined layers within the LLM (e.g., layers 0, 5, 10, ...), we dynamically insert another set of layer-specific learnable prompt vectors before the input of that layer's Transformer block.

This deep injection method allows us to exert more fine-grained guidance at different stages of information processing within the model, effectively activating task-relevant knowledge without altering the LLM's massive original parameters. The final input sequence structure for the LLM is $[\mathbf{P}_{\text{in}}; \mathbf{h}_{\text{fused}}; V_{\text{proj}}]$, which transforms into $[\mathbf{P}_{l}; \mathbf{H}_{l-1}]$ at specific layers, where $\mathbf{H}_{l-1}$ is the output from the previous layer.

\subsection{Few-shot Fault Diagnosis Fine-tuning based on Synergistic Reasoning}
\label{subsec:method_finetuning}

After establishing the dynamic synergistic prompting and deep prompt tuning mechanisms, we fine-tune the model end-to-end on a few-shot basis. This stage aims to integrate the LLM's general reasoning capabilities with domain-specific knowledge of fault diagnosis while maintaining extremely high parameter efficiency.

\subsubsection{Parameter-Efficient Fine-Tuning}
Directly fine-tuning an LLM with billions of parameters on a small dataset is not only computationally expensive but also highly prone to overfitting. To address this, we adopt LoRA. The core idea of LoRA is to freeze all the weights of the pre-trained model and inject trainable low-rank matrices in parallel with the model's key modules (such as the query, key, value, and output projection matrices of the self-attention mechanism). For a pre-trained weight matrix $W_0 \in \mathbb{R}^{d \times k}$, its update is constrained to $\Delta W = BA$, where $B \in \mathbb{R}^{d \times r}$ and $A \in \mathbb{R}^{r \times k}$ are trainable low-rank decomposition matrices, with the rank $r \ll \min(d, k)$. The model's forward pass becomes $h = (W_0 + \Delta W)x = W_0x + BAx$. By optimizing only the extremely small matrices $A$ and $B$, LoRA can achieve performance close to full fine-tuning at a cost of less than 0.1\% of the total parameters, effectively preventing the destruction of the model's original knowledge.

\subsubsection{Synergistic Reasoning and Classification}
During the fine-tuning process, the model follows a synergistic reasoning paradigm. For each training sample, the model receives its visual information along with the complete set of text prompts containing descriptions of all $C$ candidate categories. The multimodal information stream, enhanced by dynamic synergistic prompt fusion and deep prompts, is deeply processed through the LoRA-adapted LLM layers. The powerful reasoning ability of the LLM is leveraged to analyze the complex relationships between the visual features and the dynamically generated mixed-text context.
The final classification decision is based on the output representation of a specialized fusion token. This token is the final hidden state $\mathbf{h}_{\text{fusion\_out}}$ of the aforementioned $\mathbf{h}_{\text{fused}}$ after being processed by the entire LLM. We posit that this vector most fully encodes the discriminative information derived from comparing and reasoning about the image against all category texts. A simple linear classifier is attached to the output of this token, mapping it to the final logits for the $C$ classes:
\begin{equation}
    \text{Logits} = W_{\text{cls}}\mathbf{h}_{\text{fusion\_out}} + b_{\text{cls}}.
\end{equation}
The entire model, including the visual extractor, fusion module, all learnable prompts, and LoRA adapters, is jointly optimized end-to-end by minimizing the standard cross-entropy loss function $\mathcal{L}_{\text{CE}}$. This design enables the model to converge quickly under few-shot conditions and learn a robust fault diagnosis capability.

\subsection{Knowledge Distillation for Edge Deployment}
\label{subsec:method_distillation}

To deploy the powerful diagnostic capabilities of the large cloud-based multimodal model onto resource-constrained edge devices, we designed a Knowledge Distillation framework tailored for edge deployment. As illustrated in Figure~\ref{fig:placeholder}~(c), this framework aims to efficiently transfer knowledge from a Teacher Model (with hidden dimension $D_T$) to a lightweight Student Model (with hidden dimension $D_S$, where $D_S < D_T$). The core of this process lies in an innovative "Reverse Adapter" architecture and a feature alignment-based optimization strategy.

\subsubsection{Reverse Adapters and Shared Classification Head Architecture}
To maximally preserve the powerful feature extraction capabilities of the teacher model, we adopt a component reuse strategy, where the student model directly utilizes the teacher's well-trained and frozen visual extractor and gated fusion unit. To resolve the dimension mismatch between them, we design lightweight linear layers—"Reverse Adapters"—as trainable "bridges." Based on the data flow, the adapters are classified into two types.

The first type is the Post-processing Adapter, applied to the visual extractor. The teacher's visual extractor $f_{v,T}$ converts an input image $I$ into high-dimensional features $V_T \in \mathbb{R}^{N_v \times D_T}$. A post-processing adapter $f_{A_v}$ then down-projects these features to the student model's dimension:
\begin{equation}
    V_S = f_{A_v}(f_{v,T}(I)) = f_{v,T}(I) W_{A_v} + b_{A_v},
\end{equation}
where $W_{A_v} \in \mathbb{R}^{D_T \times D_S}$ is a trainable weight, and $V_S \in \mathbb{R}^{N_v \times D_S}$ is the visual sequence processable by the student model.

The second type is the Sandwich Adapter, applied to the gated fusion unit. The student's low-dimensional visual features $\mathbf{v}_{g,S}$ and text features $\mathbf{t}_{g,\text{mix},S}$ are first up-projected to $D_T$ by two independent input adapters, $f_{A_{\text{in},v}}$ and $f_{A_{\text{in},t}}$. Subsequently, the frozen teacher fusion gate $f_{\text{gate},T}$ processes the up-projected features. Finally, an output adapter $f_{A_{\text{out}}}$ down-projects the fusion result back to $D_S$. The entire process can be represented as:
\begin{equation}
    \mathbf{h}_{\text{fused},S} = f_{A_{\text{out}}}(f_{\text{gate},T}(f_{A_{\text{in},v}}(\mathbf{v}_{g,S}), f_{A_{\text{in},t}}(\mathbf{t}_{g,\text{mix},S}))).
\end{equation}
Beyond component-level adaptation, the most critical design of the entire architecture is that the student and teacher models share the same classification head $f_{\text{cls}}$. This implies that after processing by its own LLM, the student model's final feature vector $\mathbf{h}_{S} \in \mathbb{R}^{D_S}$ must be up-projected to $\mathbf{h}'_{S} \in \mathbb{R}^{D_T}$ via a main adapter layer $f_{A_{\text{main}}}$ before being fed into the classification head. This series of designs imposes a strong constraint, laying the structural foundation for subsequent efficient feature distillation.

\subsubsection{Feature Alignment-based Knowledge Distillation}
Based on the above architecture, the entire distillation process is driven by a single, efficient objective: to precisely align the student and teacher models at the feature level. We extract the fused feature vector $\mathbf{h}_{T} \in \mathbb{R}^{D_T}$ from the last layer of the teacher model, which is considered its most refined and core knowledge representation of the fault information. The student model's feature after transformation by the main adapter layer is $\mathbf{h}'_{S}$. Our optimization objective is to minimize the Mean Squared Error (MSE) between these two feature vectors, which serves as the sole distillation loss function $\mathcal{L}_{\text{distill}}$:
\begin{equation}
    \mathcal{L}_{\text{distill}} = \text{MSE}(\mathbf{h}'_{S}, \mathbf{h}_{T}), \quad \text{where } \mathbf{h}'_{S} = f_{A_{\text{main}}}(\mathbf{h}_{S}).
\end{equation}
By optimizing this objective end-to-end, the student model is guided to reproduce the teacher's internal representations. Since $\mathbf{h}_{T}$ is the very discriminative feature the teacher model uses for accurate classification, forcing $\mathbf{h}'_{S}$ to approximate $\mathbf{h}_{T}$ means the student model indirectly learns a highly structured semantic space for the shared classification head. This pure feature-alignment distillation enables the student model to inherit the teacher's discriminative ability without direct supervision from true labels, thereby creating a lightweight edge agent whose performance approaches that of the large cloud model but with far less energy consumption and a much smaller footprint.

\subsection{Cloud-Edge Collaborative Online Updating based on a Shared Decision Space}
\label{subsec:method_online_update}

The synergistic intelligence framework proposed in this paper not only aims to transfer knowledge from the cloud's large model to the edge in a one-off manner but also establishes a sustainable, iterative Cloud-Edge Collaborative Online Updating mechanism. This mechanism allows the cloud model to be updated rapidly and at near-zero cost using new data collected by the edge device in practical applications, solving the long-standing problem of large models having long and costly iteration cycles. Its theoretical foundation stems from the shared decision space constructed during the knowledge distillation phase. By forcing the student and teacher models to share a classification head and optimizing for feature alignment, we ensure that the student's feature $\mathbf{h}'_{S}$ (after the main adapter transformation) and the teacher's final feature $\mathbf{h}_{T}$ are semantically equivalent and interchangeable.

The feasibility of this online updating mechanism can be rigorously proven through gradient analysis. Our goal is to demonstrate that, based on the shared decision space, the gradient update for the classification head parameters $\theta_{\text{cls}}$ calculated using the edge-side student model is highly consistent with the gradient calculated using the cloud-side teacher model. Let the feature extractors for the teacher and student be $f_{\text{feat},T}$ and $f_{\text{feat},S}$ respectively, the main adapter be $f_{A_{\text{main}}}$, the shared classification head be $f_{\text{cls}}$, and the loss function be the cross-entropy $\mathcal{L}_{\text{CE}}$. For a new sample $(I, y)$, if we directly fine-tune the teacher model's classification head in the cloud, the gradient of its loss function $\mathcal{L}_T$ with respect to $\theta_{\text{cls}}$ is:
\begin{equation}
\nabla_{\theta_{\text{cls}}} \mathcal{L}_T = \frac{\partial \mathcal{L}_{\text{CE}}(f_{\text{cls}}(f_{\text{feat},T}(I); \theta_{\text{cls}}), y)}{\partial \theta_{\text{cls}}}.
\end{equation}
On the edge side, we update by fine-tuning the student model's classification head, and the gradient of its loss function $\mathcal{L}_S$ with respect to $\theta_{\text{cls}}$ is:
\begin{equation}
\nabla_{\theta_{\text{cls}}} \mathcal{L}_S = \frac{\partial \mathcal{L}_{\text{CE}}(f_{\text{cls}}(f_{A_{\text{main}}}(f_{\text{feat},S}(I)); \theta_{\text{cls}}), y)}{\partial \theta_{\text{cls}}}.
\end{equation}
Since our optimization objective during knowledge distillation is to minimize the feature difference, i.e., $f_{A_{\text{main}}}(f_{\text{feat},S}(I)) \approx f_{\text{feat},T}(I)$, and because the loss function $\mathcal{L}_{\text{CE}}$ and its gradient are continuous, a small difference in the input features will result in a small difference in the gradient vectors. Therefore, we can conclude that the two gradients are highly consistent in both direction and magnitude:
\begin{equation}
\nabla_{\theta_{\text{cls}}} \mathcal{L}_S \approx \nabla_{\theta_{\text{cls}}} \mathcal{L}_T.
\end{equation}
This derivation fundamentally proves the effectiveness of our online updating framework. In practice, the process manifests as an efficient closed loop: First, the lightweight student model deployed at the edge uses new data to perform a quick fine-tuning of only the classification head $f_{\text{cls}}$, which has a very small number of parameters, while keeping its large feature extraction module frozen. This yields an updated head $f'_{\text{cls}}$. Next, only the weights of $f'_{\text{cls}}$—a minimal amount of data—need to be uploaded to the cloud. Finally, the large teacher model in the cloud directly swaps in the new classification head $f'_{\text{cls}}$, completing the instantaneous knowledge synchronization. This entire process avoids any costly retraining on the cloud, enabling the continuous evolution of the large cloud model driven by the low computational consumption at the edge, thus providing a new paradigm for building truly dynamic and adaptive industrial intelligence systems.

%% file: Experiment_seting.tex
\begin{figure}
    \centering
    % 第一行两张图
    \begin{subfigure}[b]{0.45\textwidth}
        \centering
        \includegraphics[width=\textwidth]{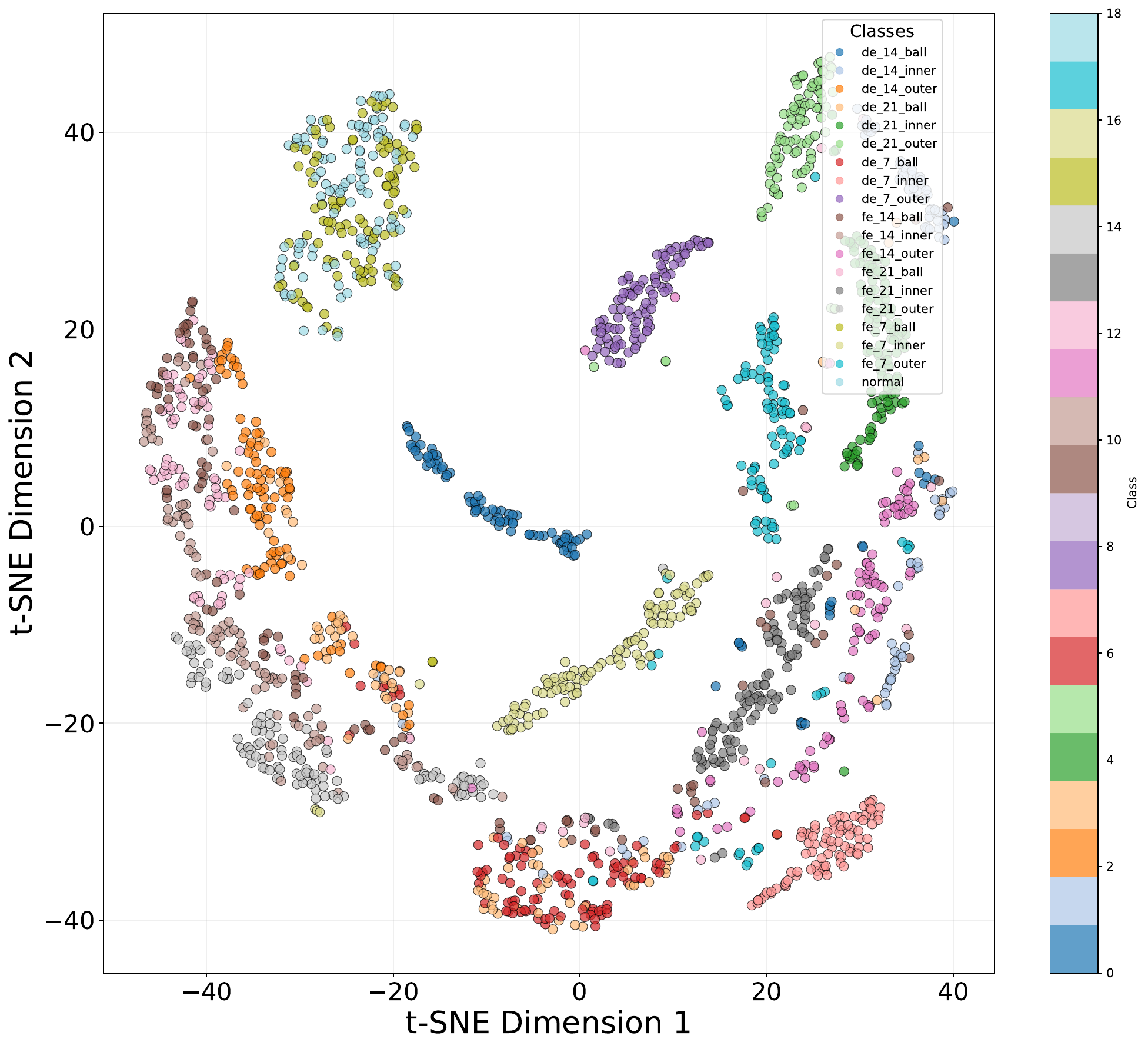}
        \caption{All features (no alignment)}
        \label{fig:CWRU_SNE_1}
    \end{subfigure}
    \hspace{0.1cm} % 调整此处的间距值
    \begin{subfigure}[b]{0.45\textwidth}
        \centering
        \includegraphics[width=\textwidth]{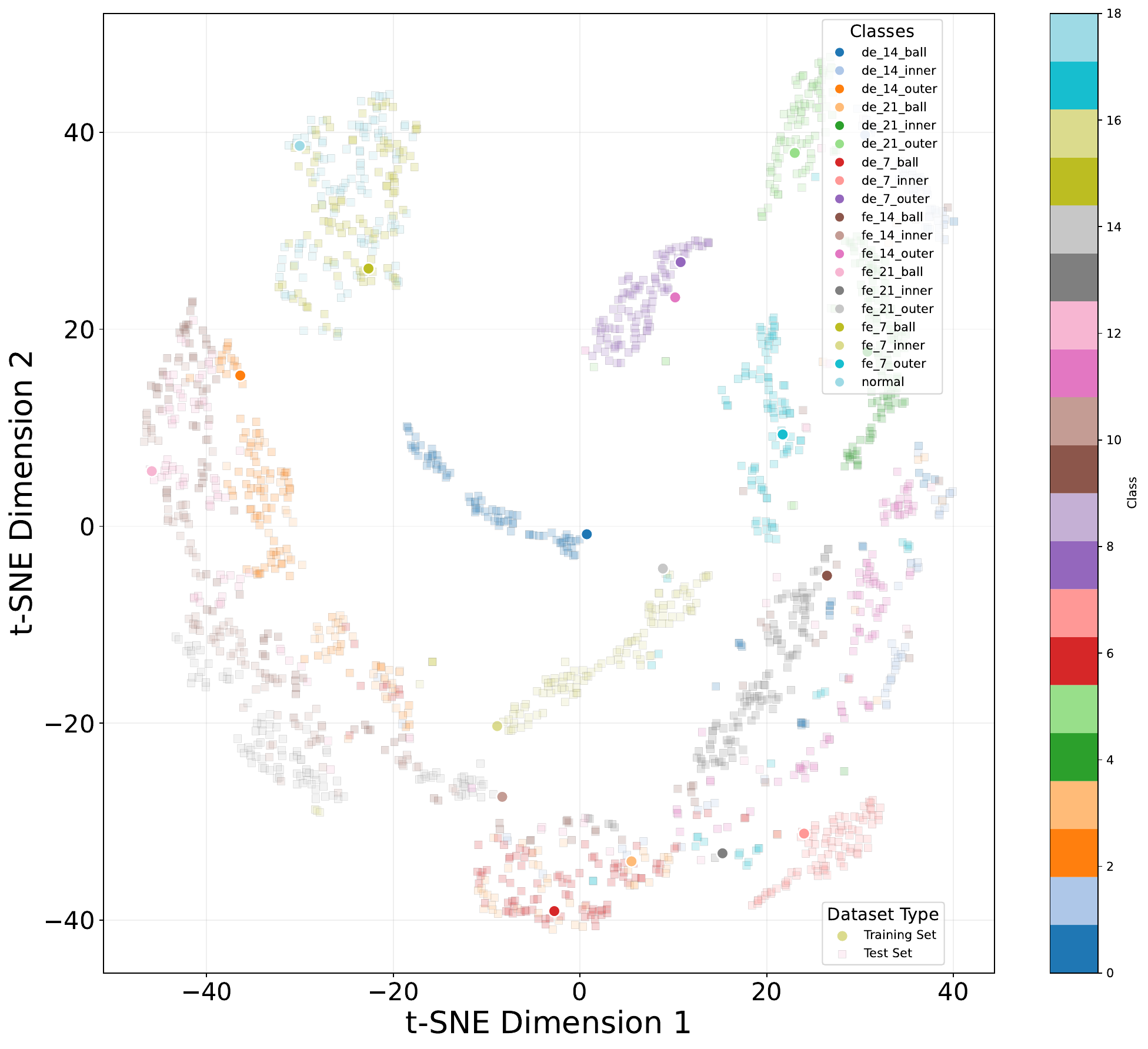}
        \caption{Training data vs test data (no alignment)}
        \label{fig:CWRU_SNE_2}
    \end{subfigure}
    
    \vspace{0.1cm} % 两行之间的间距
    
    % 第二行两张图
    \begin{subfigure}[b]{0.45\textwidth}
        \centering
        \includegraphics[width=\textwidth]{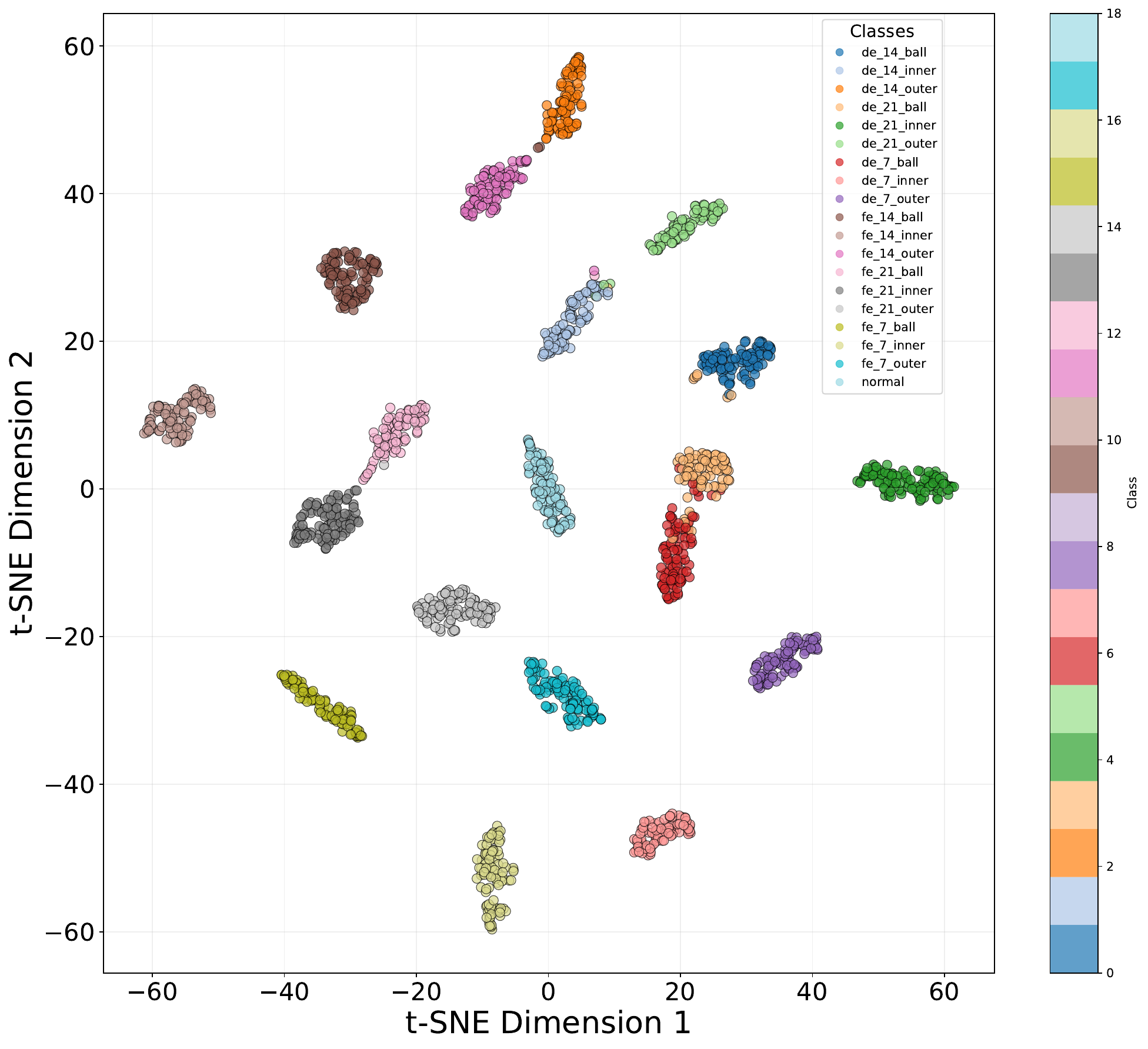}
        \caption{All features (after alignment)}
        \label{fig:CWRU_SNE_3}
    \end{subfigure}
    \hspace{0.1cm} % 调整此处的间距值
    \begin{subfigure}[b]{0.45\textwidth}
        \centering
        \includegraphics[width=\textwidth]{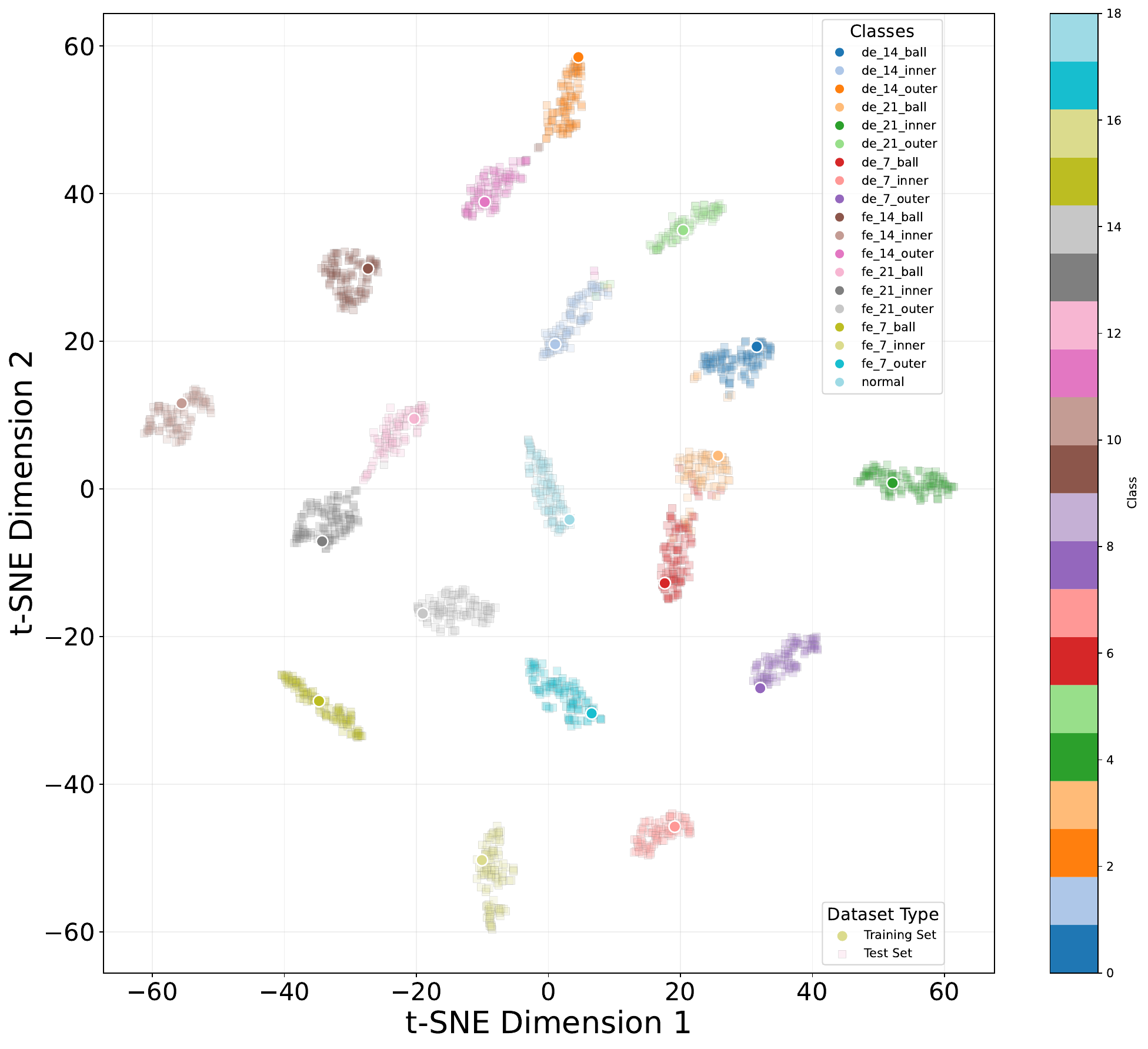}
        \caption{Training data vs test data (after alignment)}
        \label{fig:CWRU_SNE_4}
    \end{subfigure}
    
    \caption{Feature distribution using t-SNE on the CWRU dataset: before and after image-text alignment}
    \label{fig:CWRU_SNE}
\end{figure}

\begin{figure}
    \centering
    % 第一行两张图
    \begin{subfigure}[b]{0.45\textwidth}
        \centering
        \includegraphics[width=\textwidth]{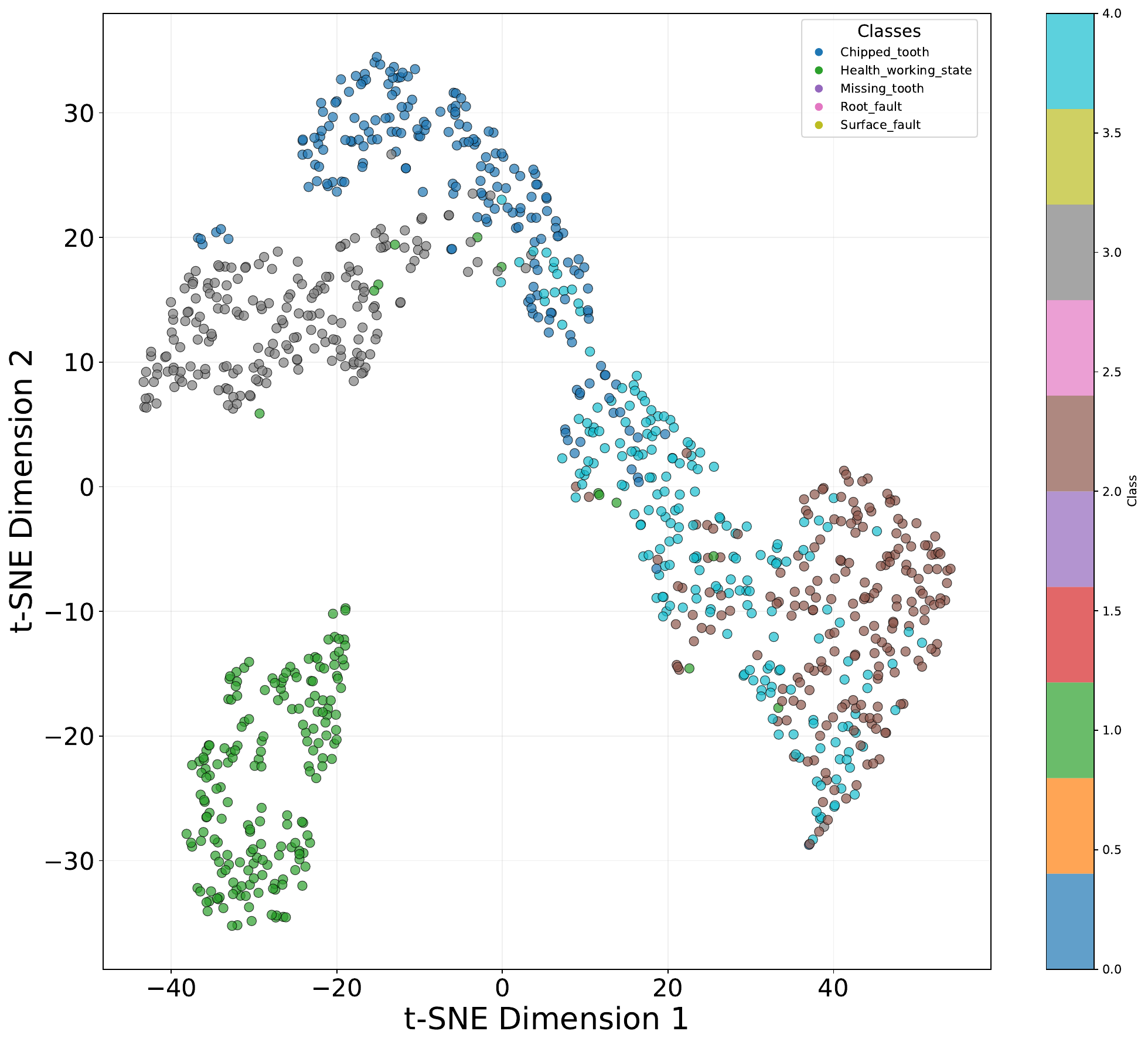}
        \caption{All Features (no alignment)}
        \label{fig:SEU_SNE_1}
    \end{subfigure}
    \hspace{0.1cm} % 调整此处的间距值
    \begin{subfigure}[b]{0.45\textwidth}
        \centering
        \includegraphics[width=\textwidth]{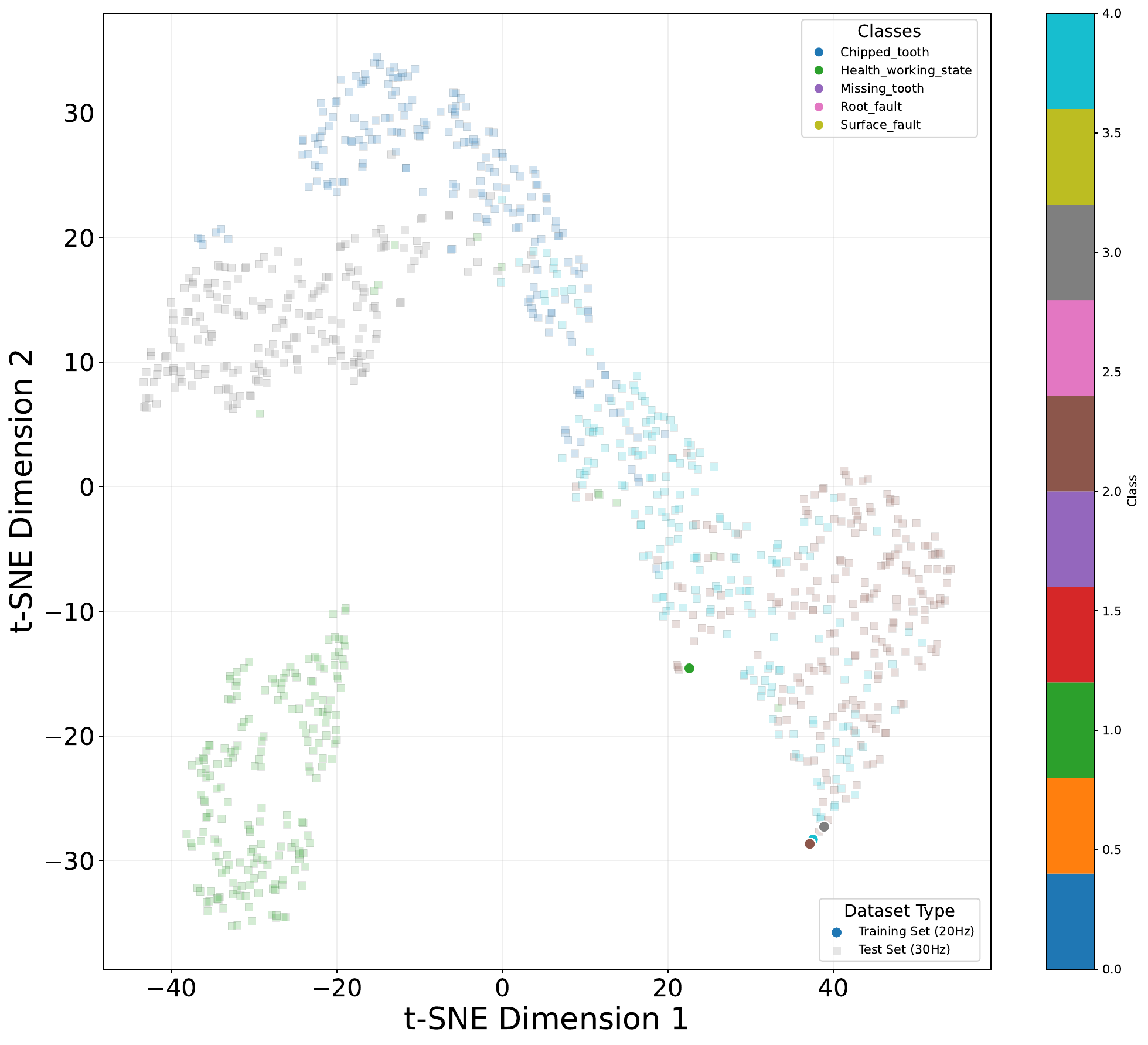}
        \caption{Train vs Test (no alignment)}
        \label{fig:SEU_SNE_2}
    \end{subfigure}
    
    \vspace{0.1cm} % 两行之间的间距
    
    % 第二行两张图
    \begin{subfigure}[b]{0.45\textwidth}
        \centering
        \includegraphics[width=\textwidth]{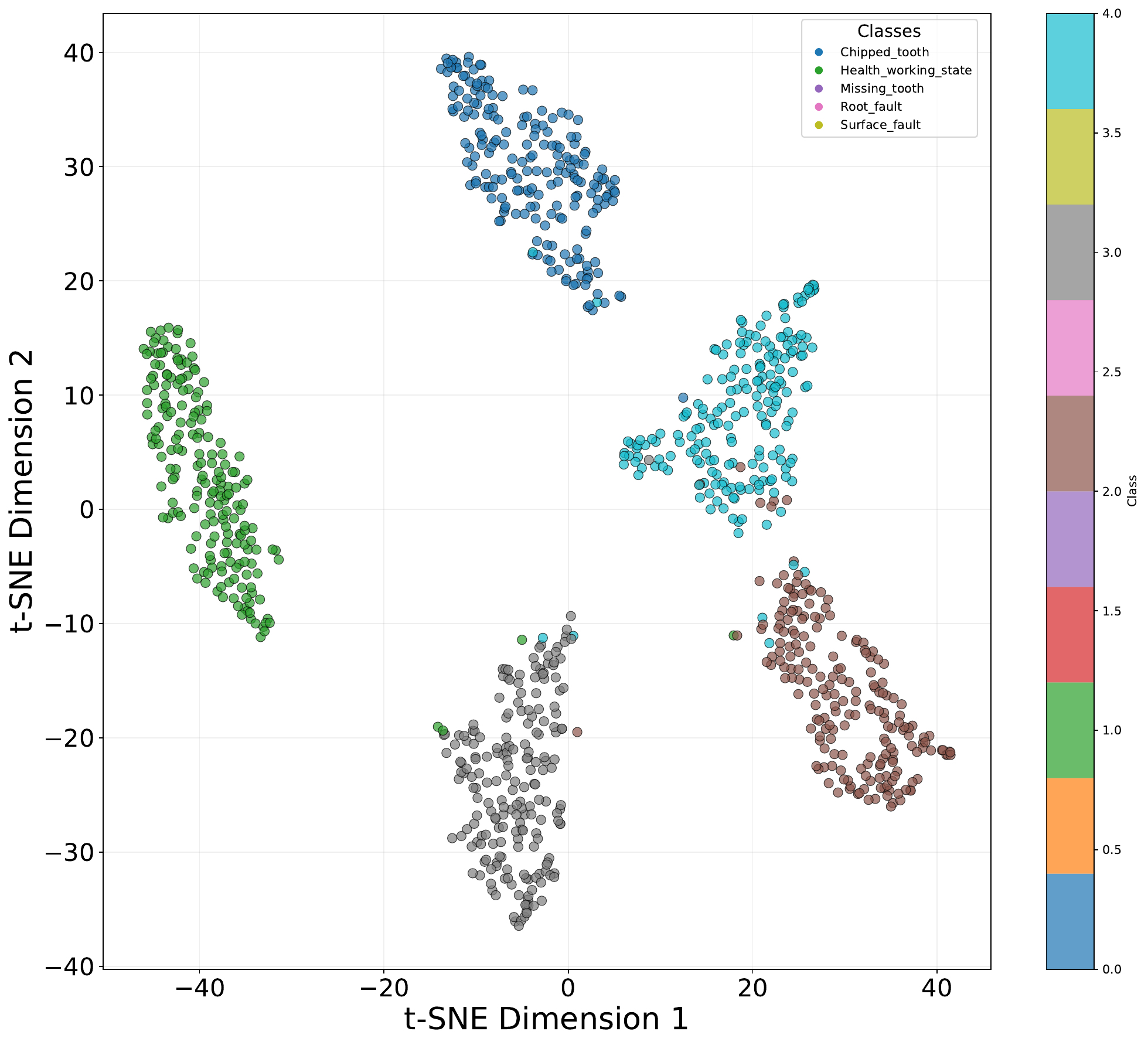}
        \caption{All Features (with alignment)}
        \label{fig:SEU_SNE_3}
    \end{subfigure}
    \hspace{0.1cm} % 调整此处的间距值
    \begin{subfigure}[b]{0.45\textwidth}
        \centering
        \includegraphics[width=\textwidth]{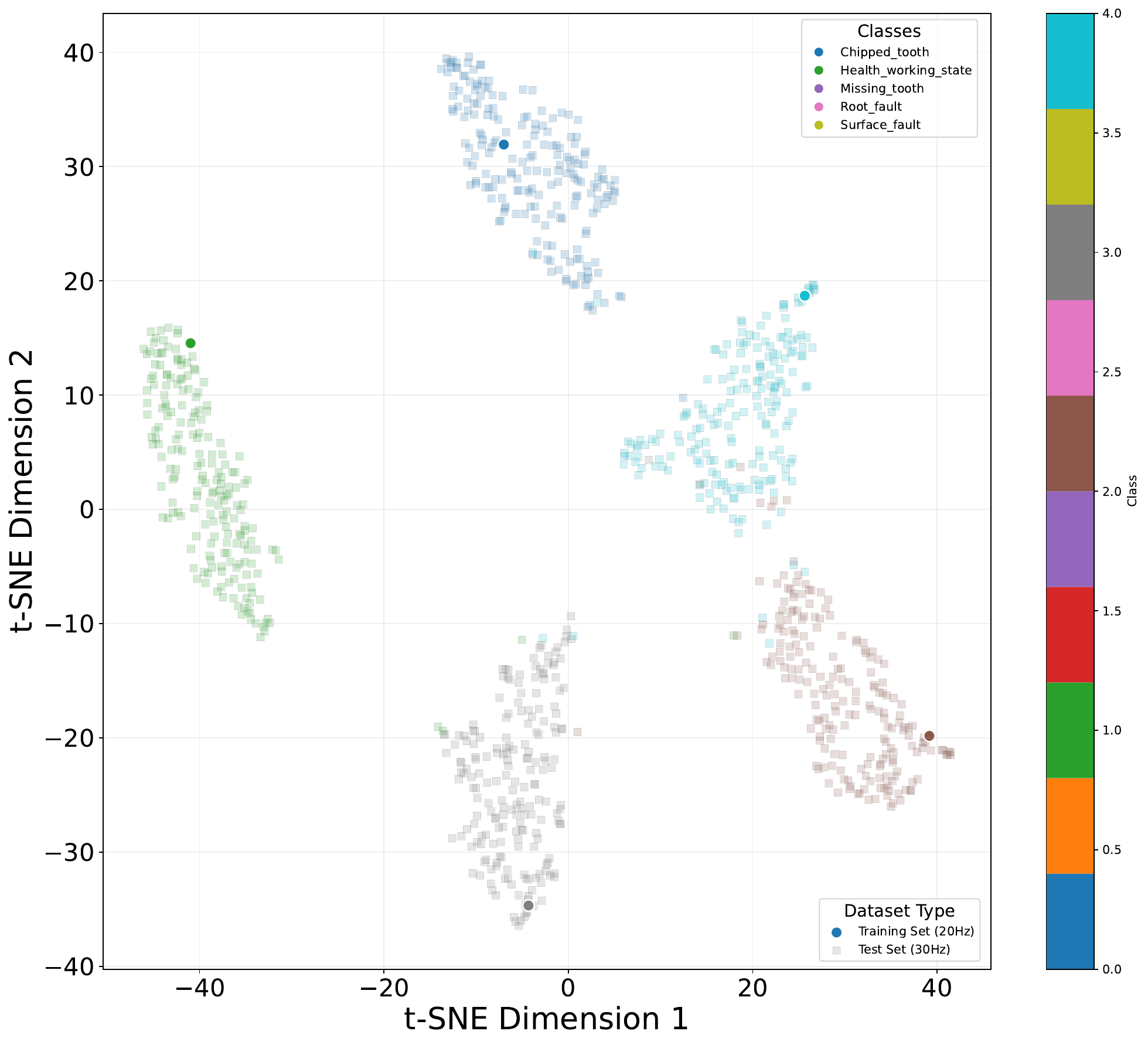}
        \caption{Train vs Test (with alignment)}
        \label{fig:SEU_SNE_4}
    \end{subfigure}
    
    \caption{Feature distribution using t-SNE on the SEU dataset: before and after image-text alignment}
    \label{fig:SEU_SNE}
\end{figure}

\section{Experimental setup}

\subsection{Dataset}

Our experiments are primarily based on two widely used public datasets in the field of fault diagnosis. For each dataset, we pair the sensor data with a corresponding textual description of the machine's health state.
Case Western Reserve University (CWRU) Bearing Dataset~\cite{smith2015rolling}:
This is a standard benchmark dataset for rotating machinery fault diagnosis. To comprehensively evaluate the generalization and robustness of our model, we used data from four different operating conditions (corresponding to motor speeds of 1730, 1750, 1772, and 1797 RPM). For each operating condition, we used 19 categories. These include one normal state, described with the text prompt "normal operation", and 18 distinct fault types. Each fault is explicitly described by its location (motor end or fan end), type (inner race, ball, or outer race), and damage diameter (0.007, 0.014, or 0.021 inches). For example, a specific fault is described with the text prompt "inner race fault at motor end with diameter 0.007 inches."

Southeast University (SEU) Gearbox Dataset~\cite{shao2018highly}:
This dataset contains more complex gearbox faults. We selected data from two different operating conditions (20Hz-0V and 30Hz-2V). This dataset covers five gearbox  fault types, each associated with a descriptive text prompt: chipped tooth ("Chipped tooth fault"), healthy working state ("Healthy working state"), missing tooth("Missing tooth fault"), root fault ("Root fault"), and surface fault ("Surface fault").

For both datasets, raw vibration signals were first preprocessed via sliding window (window size: 512, overlap ratio: 0.5, window step: 256), then converted to time-frequency spectra using continuous wavelet transform (CWT) — with the complex Morlet wavelet as the basis function, total scale of 128, sampling period (reciprocal of sampling frequency), and 200 samples uniformly selected per category to balance classes. Both datasets adopted identical sliding window and CWT settings, and the resulting spectra served as the model’s visual input, paired with corresponding text prompts.

\subsection{Implementation details}

Our models, Syn-Diag-Cloud (LLaMA-3.1-8B-Instruct) and Syn-Diag-Edge (LLaMA-3.1-1B), were trained on an NVIDIA A40 GPU using the AdamW optimizer for all stages. The training set was constructed in an N-way K-shot format, totaling N×K samples. For the CWRU and SEU datasets, the test set for each working condition was fixed at 1834 and 480 samples, respectively.
During the cross-modal pre-training phase, we froze the LLM and trained only the visual feature extractor for 600 (CWRU) and 1000 (SEU) epochs with a learning rate of $1 \times 10^{-2}$.
For the few-shot fine-tuning stage, we employed LoRA (rank=16, alpha=32) for 20 epochs. This stage and the subsequent knowledge distillation stage, which used an MSE loss for 40 epochs, shared a learning rate of $3 \times 10^{-4}$ and a batch size of 4. A weight decay of 0.01 was also applied during fine-tuning.
To validate the online updating mechanism, we simulated 30 update cycles. In each cycle, only the student model's classification head was fine-tuned for 10 epochs on 3 new samples per class, using a learning rate of $1 \times 10^{-4}$ and a batch size of 16.

%% file: Experiment.tex
 \input{table1}

\section{Experimental Results}

\subsection{Validation of Cross-Modal Semantic Alignment Effectiveness}

This experiment aims to qualitatively and quantitatively validate the effectiveness of our proposed multi-level cross-modal semantic alignment pre-training method. The core objective is to evaluate whether the visual feature extractor, after being trained in this stage, can generate a feature space with high semantic separability.

To obtain general fault representations that can span different working conditions, we first pre-train the visual extractor with cross-modal semantic alignment. The training for this stage is conducted independently for different datasets: on the CWRU dataset, we pool data from all four of its working conditions; on the SEU dataset, we pool data from its two working conditions. During their respective pre-training processes, we extract 7 labeled samples (i.e., 7-shot) for each fault category to train the corresponding visual extractor. To intuitively assess the quality of the learned features, we use the t-SNE dimensionality reduction technique to visualize the visual feature distributions on both the CWRU and SEU datasets before and after pre-training. The experimental results are shown in Figure~\ref{fig:CWRU_SNE} and Figure~\ref{fig:SEU_SNE}. Before alignment pre-training, the features of samples from different fault categories are severely mixed in the 2D space, failing to form any meaningful cluster structure. This indicates that the original, unaligned visual feature space lacks semantic discriminative ability, and the model cannot effectively distinguish different fault patterns.

In stark contrast, after our semantic alignment pre-training, the feature distribution undergoes a fundamental improvement. As shown in Figure~\ref{fig:CWRU_SNE_3} and Figure~\ref{fig:SEU_SNE_3}, samples of the same fault category form highly compact and clearly bounded clusters in the feature space, while the distances between different categories are significantly increased. This phenomenon strongly demonstrates that our method successfully injects the semantic knowledge of the LLM into the visual features, creating a strong correlation between the visual representations and the semantic concepts of the fault categories. More importantly, as shown in Figure~\ref{fig:CWRU_SNE_4} and Figure~\ref{fig:SEU_SNE_4}, samples from the unseen test set can accurately fall into the corresponding category clusters defined by the training set samples. This clearly indicates that our model has learned a powerful generalization capability, rather than merely memorizing the training data.

In summary, the t-SNE visualization results unequivocally validate the effectiveness of our cross-modal semantic alignment method. This method successfully reshapes the original, chaotic visual feature space into a structured, semantically rich, and high-quality feature space with excellent generalization. This lays a solid foundation for the subsequent model to achieve accurate and robust fault diagnosis under extremely few-shot conditions.

\input{table4}

\subsection{Few-shot Fault Diagnosis Performance Comparison}
\label{sec:2}

This experiment is central to evaluating the performance of our synergistic intelligence framework. Its main purpose is to verify the diagnostic capability of our proposed models (including the cloud version, Syn-Diag-Cloud, and the edge version, Syn-Diag-Edge) when faced with the common problem of scarce labeled data in industrial scenarios. We conduct experiments under strict few-shot conditions and compare the performance with several baseline models.

We follow the standard few-shot learning paradigm, conducting $N$-way $K$-shot experiments on the CWRU and SEU datasets. Specifically, we randomly sample a very small number of samples ($K$=1, 3, 5, 7) from each fault category as the training set and use an independent, sample-rich test set to evaluate the model's final performance. To ensure fair results, all methods are trained and tested on the same data splits. We select three representative deep learning models as baselines for comparison, including a classic lightweight Convolutional Neural Network (CNN), a Deep Residual Network (ResNet-50) which excels in image recognition, and a modern visual backbone based on the self-attention mechanism, the Vision Transformer (ViT). To ensure a fair comparison and to demonstrate that the superiority of our framework stems from its synergistic mechanism rather than data volume, all baseline models also adopt a similar two-stage training paradigm. Specifically, each baseline model is first pre-trained on the exact same multi-condition 7-shot dataset as our visual extractor. Subsequently, these pre-trained baseline models, along with our models, are trained and evaluated on the final $K$-shot ($K$=1, 3, 5, 7) tasks. Our proposed Syn-Diag-Cloud model undergoes the full fine-tuning process described in this paper, while the Syn-Diag-Edge model is a lightweight version obtained through knowledge distillation from the Cloud version.

The detailed comparative results of the few-shot fault diagnosis are shown in Table~\ref{tab:cwru_results} and Table~\ref{tab:seu_results}. The results clearly indicate that under all working conditions and all few-shot settings on both the CWRU and SEU datasets, our proposed Syn-Diag-Cloud and Syn-Diag-Edge models significantly outperform all baseline methods across three key metrics: Accuracy (ACC), Precision, and F1-Score. This performance advantage is particularly prominent under the extreme condition of only one labeled sample (1-shot). For example, under the CWRU 1730 RPM condition (Table~\ref{tab:cwru_results}), Syn-Diag-Cloud achieves an astonishing accuracy of 98.09\%, whereas the accuracies for CNN, ResNet-50, and ViT are only 78.60\%, 68.84\%, and 55.39\%, respectively. This demonstrates the extremely high data efficiency of our model, enabled by the LLM's prior knowledge and the synergistic reasoning mechanism. Furthermore, our model exhibits excellent performance saturation and stability; it reaches a very high performance level even in the 1-shot condition, and as the number of samples increases, the performance improves steadily and tends to saturate, indicating a much lower dependency on the number of training samples compared to traditional methods. It is noteworthy that the Syn-Diag-Edge model, obtained through knowledge distillation, shows no significant performance degradation despite having far fewer parameters than the cloud version. In several test scenarios, its performance is on par with or even slightly exceeds that of Syn-Diag-Cloud, and it consistently outperforms all baseline models. This fully validates the effectiveness of our knowledge distillation framework in efficiently compressing the complex knowledge of a large model into a lightweight edge model.

The experimental results strongly prove the great potential of our synergistic intelligence framework in solving the problem of few-shot industrial fault diagnosis. Traditional visual models struggle to learn robust features when data is scarce, leading to poor performance. In contrast, our method (1) constructs a high-quality feature space through cross-modal semantic alignment; (2) enables the model to reason within a context that includes all possibilities through dynamic synergistic prompt fusion, significantly enhancing its discriminative power; and (3) fully leverages the LLM's powerful prior knowledge with parameter-efficient fine-tuning techniques. These three synergistic mechanisms work in concert, enabling the model to achieve accurate identification even after "seeing" a fault sample only once. The excellent performance of Syn-Diag-Edge further proves the deployability of the framework, providing a reliable path for implementing high-performance, low-cost intelligent diagnosis systems in real industrial environments.

\input{table3}

\subsection{Cross-Working-Condition Generalization Capability Evaluation}
\label{sec:3}
In real industrial environments, the working conditions of equipment (such as speed and load) are dynamic. A robust diagnostic model should not only perform well under the single working condition of the training data but also possess the ability to transfer and generalize across different conditions. This experiment aims to rigorously evaluate the performance of our proposed Syn-Diag-Cloud model when the training and testing data come from different working conditions.

We designed a comprehensive cross-condition test based on the four different working conditions of the CWRU dataset (1730, 1750, 1772, and 1797 RPM). The experiment follows a 19-way 5-shot setting, where the model is fine-tuned on a 5-shot training set from one source condition and then directly evaluated on the full test set of another completely different, unseen target condition. We tested all 12 possible source-target condition pairings to thoroughly examine the model's generalization capability.

The detailed results of the cross-condition evaluation are shown in Table~\ref{tab:cross_domain_results}. The results are extremely compelling; in all 12 cross-condition transfer tasks, our model demonstrated excellent and highly stable diagnostic performance. The test accuracy (ACC), precision, and F1-score in all scenarios remained above 95.9\%, with the accuracy in the 1772R to 1750R transfer task reaching as high as 98.58\%.

This result strongly proves that the fault knowledge learned by our model is not limited to superficial signal features of a specific working condition but has successfully captured more essential and universal fault patterns. Even when a domain shift exists between the training and testing data due to different speeds and loads, the model can still accurately identify the fault type. This powerful zero-shot cross-condition generalization ability is mainly attributed to the core advantage of our framework: by aligning visual features with stable, condition-independent natural language semantic descriptions from a large language model, the model is guided to learn the intrinsic physical meaning of the faults, rather than the data distribution of a specific condition. For example, regardless of how the speed changes, the semantic concept of "an inner race fault of 0.007 inches" is constant, and our model anchors on this constant semantic-visual mapping relationship. This experiment fully demonstrates the robustness of our model in the face of variable industrial environments, providing solid data support for its reliable deployment in practical scenarios.

\begin{figure}
    \centering % 使整个图片环境居中
    
    % 第一行图片
    \begin{subfigure}[b]{0.45\textwidth} % 每张图片占页面宽度的45%
        \centering
        \includegraphics[width=\textwidth]{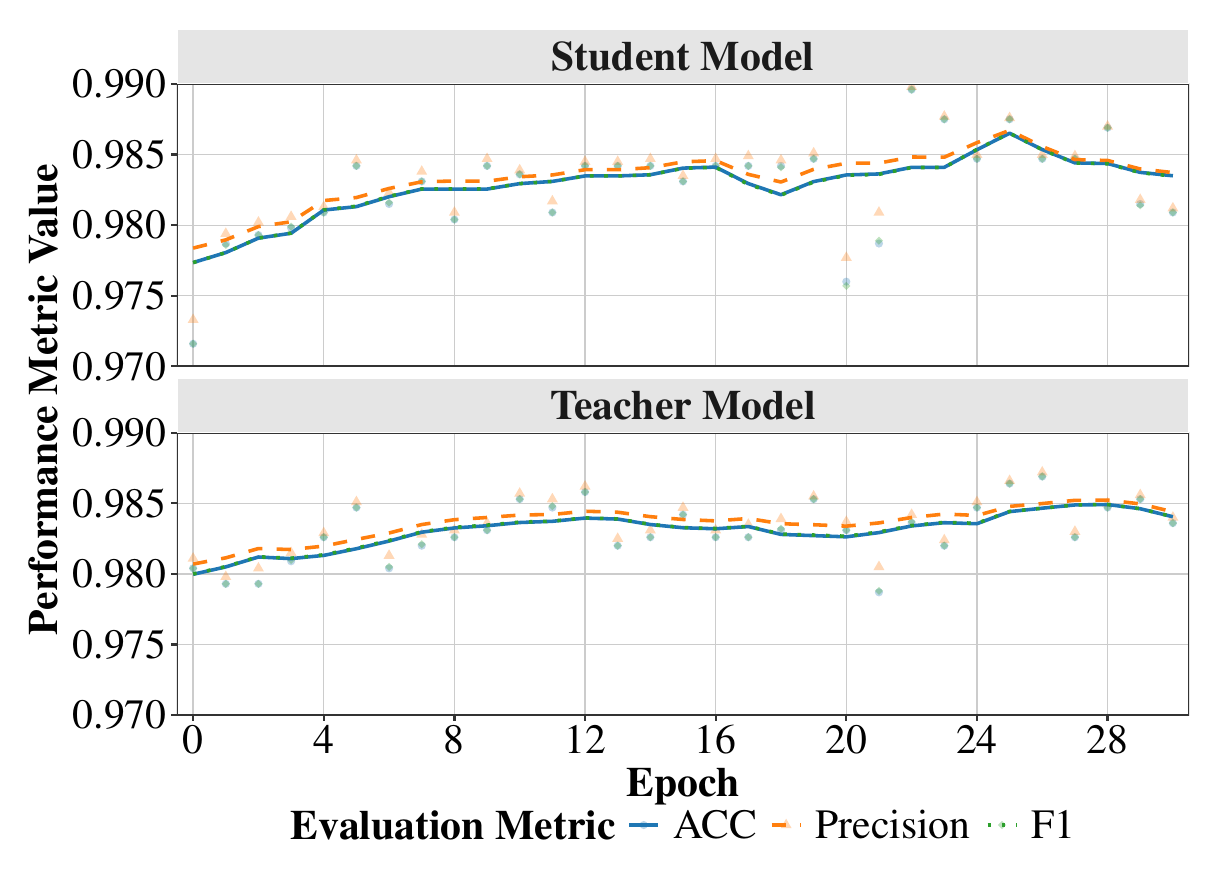} % 插入第一张图片
        \caption{1730R} % 第一张图片的标题
        \label{fig:sub1} % 第一张图片的标签，用于交叉引用
    \end{subfigure}
    \hspace{0.1cm} % 调整此处的间距值
    \begin{subfigure}[b]{0.45\textwidth}
        \centering
        \includegraphics[width=\textwidth]{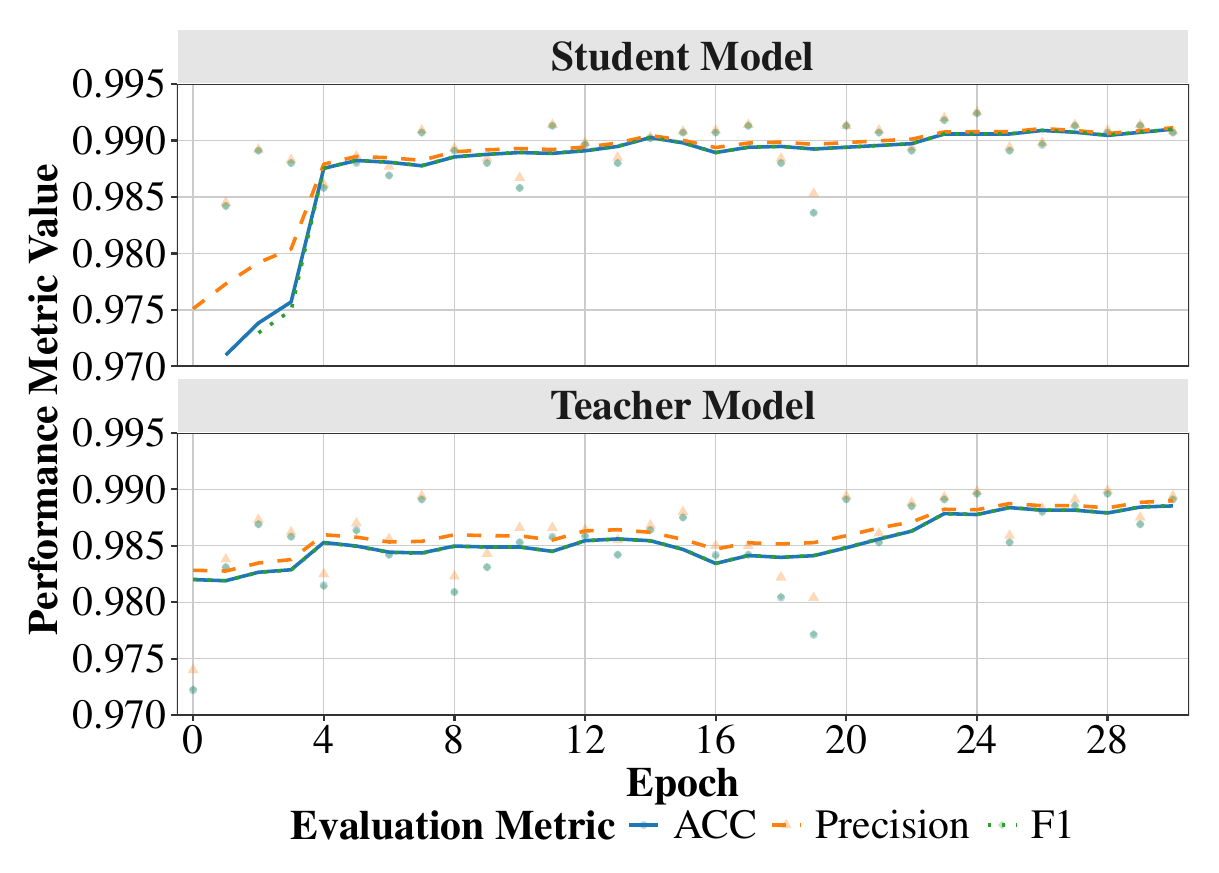} % 插入第二张图片
        \caption{1750R}
        \label{fig:sub2}
    \end{subfigure}
    
    % 第二行图片
    \begin{subfigure}[b]{0.45\textwidth}
        \centering
        \includegraphics[width=\textwidth]{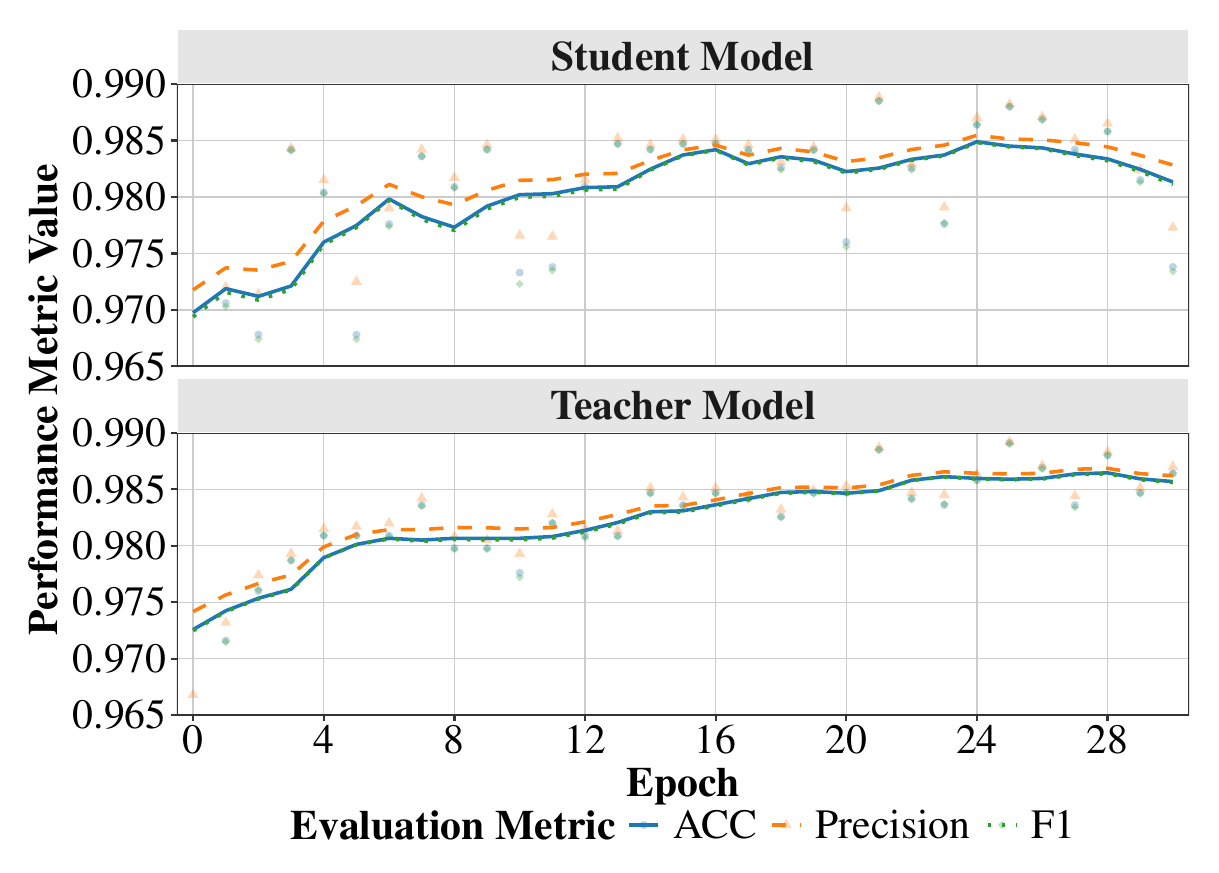} % 插入第三张图片
        \caption{1772R}
        \label{fig:sub3}
    \end{subfigure}
    \hspace{0.1cm} % 调整此处的间距值
    \begin{subfigure}[b]{0.45\textwidth}
        \centering
        \includegraphics[width=\textwidth]{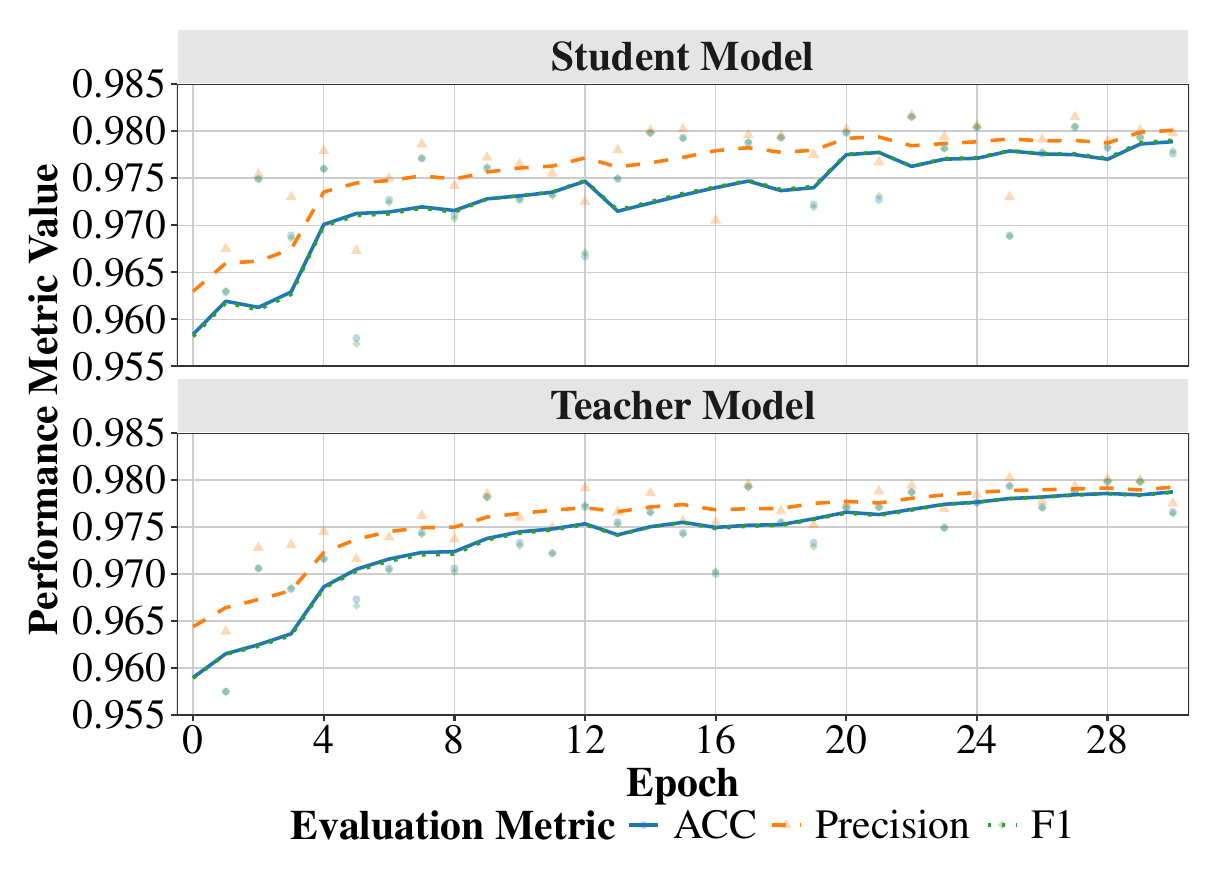} % 插入第四张图片
        \caption{1797R}
        \label{fig:sub4}
    \end{subfigure}
    
    \caption{Effects of edge updates and cloud aggregation under different conditions on CWRU data (the teacher model here refers to Syn-Diag-Cloud, and the student model here refers to Syn-Diag-Edge)} % 整个图片组的总标题
    \label{fig:online_update_results} % 整个图片组的标签
\end{figure}

\subsection{Cloud-Edge Collaborative Online Update Performance Validation}
\label{sec:4}

To validate the practical effectiveness of the "Cloud-Edge Collaborative Online Updating based on a Shared Decision Space" mechanism, we designed a continuous learning experiment that simulates a real industrial scenario. This experiment aims to demonstrate that by fine-tuning the classification head of the student model (Syn-Diag-Edge) on the edge side with a small amount of new data and synchronizing the updated weights to the cloud, the performance of the cloud teacher model (Syn-Diag-Cloud) can be effectively improved.

We used the four working conditions of the CWRU dataset as independent test scenarios. The experiment simulates a continuous learning process: in the initial state, neither the teacher nor the student model has seen the data used for online updates. We iterate in units of cycles. At the beginning of each cycle, we randomly sample a very small number of new samples (3 per class in this experiment) from the validation set to simulate incremental data collected at the edge. Subsequently, we only fine-tune the classification head of the student model, while the rest of the model remains completely frozen. After each cycle, we evaluate two key metrics: 1) the performance of the updated student model itself; and 2) the performance of the teacher model, where the teacher model's classification head is directly replaced with the student's updated one.

The results of the online update experiment are shown in Figure~\ref{fig:online_update_results}, which clearly illustrates the evolution of model performance as the online update cycles increase. Across all four working conditions, we observe a highly consistent phenomenon: as a small amount of new data is continuously added, the performance metrics (accuracy, precision, F1-score) of both the Student Model and the Teacher Model updated by it show a steady and gradual upward trend.

This result strongly validates the effectiveness of our online update framework. Although the model performance is already at a high level initially, through very low-cost fine-tuning of the classification head at the edge and synchronizing the learned new knowledge back to the cloud, the performance of the entire system is continuously optimized and improved. It is noteworthy that the smooth rise of the teacher model's performance curve indicates that the classification head weights transferred from the edge can seamlessly integrate with the powerful feature extraction capabilities of the cloud model, effectively absorbing information from new data without causing performance oscillations or degradation. This is thanks to the shared decision space we constructed through knowledge distillation, which ensures semantic consistency at the decision level between the cloud and edge models. This experiment fully demonstrates that our synergistic intelligence framework is not just a static diagnostic tool, but a system with dynamic learning and adaptive evolution capabilities, able to iterate continuously in practical applications to maintain optimal diagnostic performance.

\begin{figure}
    \centering
    \includegraphics[width=0.9\linewidth]{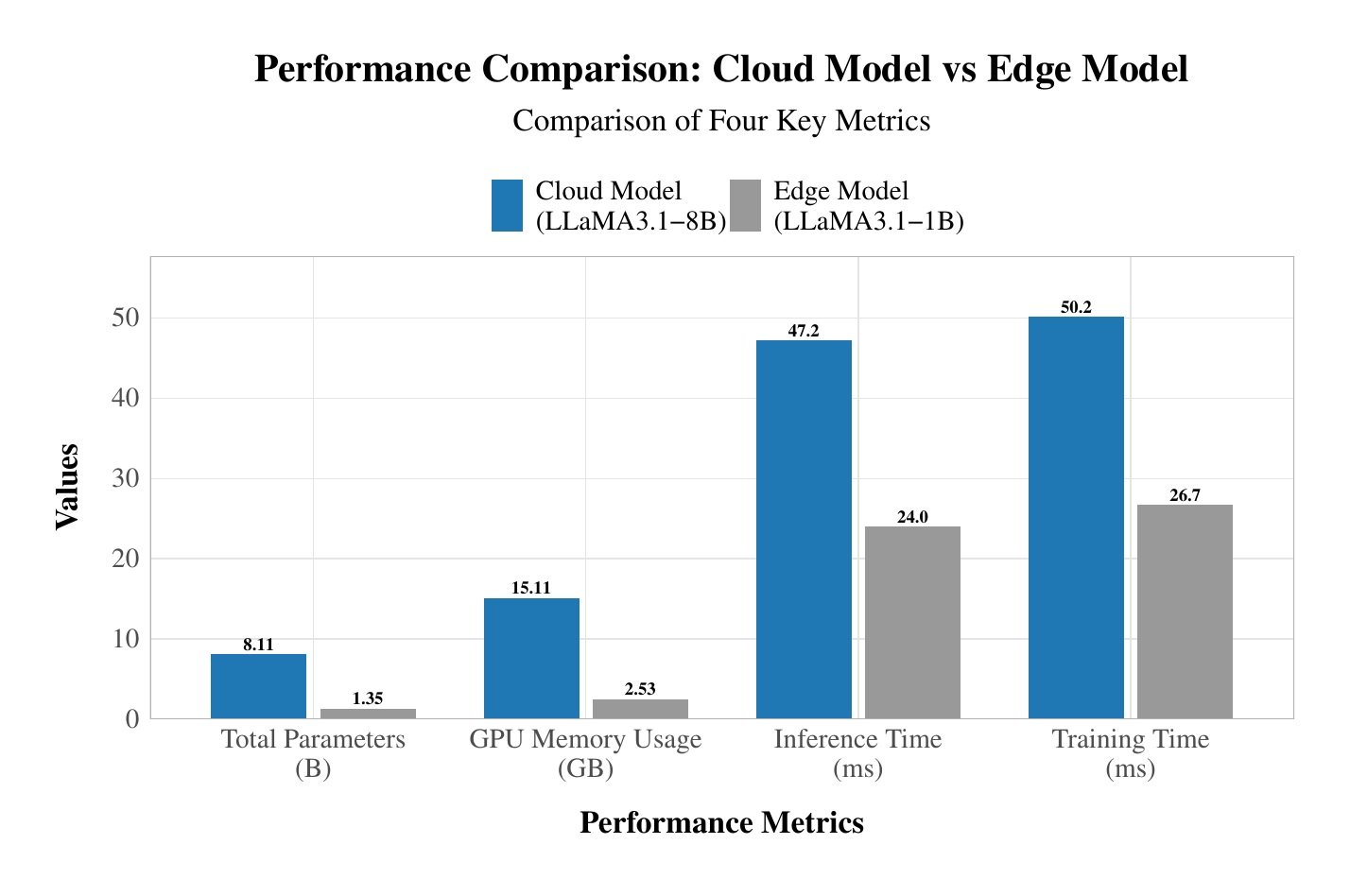}
    \caption{Resource Occupation and Efficiency Test for Syn-Diag-Cloud and Syn-Diag-Edge}
    \label{fig:efficiency_comparison}
\end{figure}
\subsection{Cloud-Edge Model Resource Consumption and Efficiency Evaluation}

After demonstrating the superiority of our framework in diagnostic accuracy, this experiment aims to quantitatively compare the cloud model (Syn-Diag-Cloud) and the knowledge-distilled edge model (Syn-Diag-Edge) in terms of computational resource consumption and operational efficiency, from a deployment feasibility perspective. This is crucial for evaluating the practical value of our cloud-edge collaborative framework in real industrial environments.

On the same hardware platform, we tested two models, Syn-Diag-Cloud (based on LLaMA3.1-8B) and Syn-Diag-Edge (based on LLaMA3.1-1B), on four key metrics: 1) Total Parameters, measuring the model size; 2) GPU Memory Usage, assessing the hardware resource requirements during runtime; 3) Inference Time per sample, evaluating the model's diagnostic response speed; and 4) Training Time per step, evaluating the efficiency of the model during online updates.

The resource and efficiency comparison results for the cloud-edge models are shown in Figure~\ref{fig:efficiency_comparison}. The experimental data clearly demonstrate the significant success of our knowledge distillation and model compression efforts. In terms of model size, Syn-Diag-Edge has only 1.35B parameters, achieving a compression of about 6 times compared to Syn-Diag-Cloud's 8.11B. This substantial reduction in scale directly leads to a significant decrease in resource consumption, with GPU memory usage dropping from 15.11 GB to 2.53 GB, a reduction of nearly 83\%. This change eliminates the dependency on expensive data-center-grade hardware, allowing the model to be easily deployed on industrial edge computing devices with limited GPU memory.

In terms of operational efficiency, the advantages of the edge model are equally apparent. Its average inference time per sample is 24.0ms, nearly twice as fast as the cloud model's 47.2ms, meeting the demand for real-time or near-real-time diagnosis in industrial scenarios. Meanwhile, the training time per step is also reduced from 50.2ms to 26.7ms, which means that the edge can iterate with new data more efficiently when performing online update tasks.

In conclusion, the results of this experiment, together with the results from sections~\ref{sec:2}, \ref{sec:3}, and \ref{sec:4}, form a complete loop: we have not only achieved state-of-the-art (SOTA) performance that surpasses existing methods in accuracy but have also, through the cloud-edge collaborative framework, successfully presented this high performance in a lightweight, efficient, and low-cost form, making it ready for large-scale deployment in real industrial environments. This strongly proves that our proposed synergistic intelligence framework successfully addresses the dual challenges of "data scarcity" and "deployment difficulty" mentioned in the introduction.

%% file: table1.tex
\begin{table}[!t]
\caption{The training and test data on the CWRU dataset come from the same fault diagnosis analysis evaluation for the 19-way task under the same working conditions (1-shot/3-shot/5-shot/7-shot are distinguished by row, \textbf{bold} indicates the best result, \underline{underline} indicates the suboptimal result)}
\label{tab:cwru_results}
\centering
\scalebox{0.7}{ % 列数不变，适当放大比例
    \begin{tabular}{lcccccccccccc}
        \toprule
        & \multicolumn{3}{c}{1730R} & \multicolumn{3}{c}{1750R} & \multicolumn{3}{c}{1772R} & \multicolumn{3}{c}{1797R} \\
        \cmidrule(lr){2-4} \cmidrule(lr){5-7} \cmidrule(lr){8-10} \cmidrule(lr){11-13}
        Methods    & ACC   & Precision      & F1     & ACC  & Precision   & F1      & ACC     & Precision  & F1 & ACC & Precision & F1 \\
        \midrule
        % CNN系列（不同shot）
        CNN-1shot        & 0.7860 & 0.8166 & 0.7790  & 0.8308 & 0.8376 & 0.8313  & 0.7193 & 0.7523 & 0.7179 & 0.7246 & 0.7401  & 0.7214 \\
        CNN-3shot         & 0.8591 & 0.8674 & 0.8583  & 0.8074 & 0.8096 & 0.8044  & 0.7684 & 0.7913 & 0.7620 & 0.7876 & 0.8076  & 0.7840 \\
        CNN-5shot         & 0.8591 & 0.8688 & 0.8582  & 0.8250 & 0.8363 & 0.8202  & 0.8362 & 0.8417 & 0.8326 & 0.7395 & 0.7678 & 0.7320 \\
        CNN-7shot         & 0.8906 & 0.8981 & 0.8874  & 0.8767 & 0.8866 & 0.8754  & 0.8335 & 0.8577 & 0.8328 & 0.8447 & 0.8566  & 0.8457 \\
        
        \midrule
        % Transformer系列
        ViT-1shot     & 0.5539 & 0.6149 & 0.5500  & 0.3367 & 0.3966 & 0.3085  & 0.5245 & 0.5402 & 0.5099 & 0.5181 &  0.5586 & 0.5175 \\
        ViT-3shot    & 0.6297 & 0.6748 & 0.6077 & 0.4605 & 0.4532 & 0.4271  & 0.5523 & 0.5656 & 0.5283 &0.5672 & 0.5829 & 0.5619 \\
        ViT-5shot     & 0.5907 & 0.6486 & 0.5678 & 0.5608 & 0.5913 & 0.5332  & 0.5112 & 0.5484 & 0.4814 & 0.5378 & 0.6068 & 0.5410 \\
        ViT-7shot     & 0.6638 & 0.7126 & 0.6646 & 0.5640 & 0.6119 & 0.5543  & 0.6270 & 0.6432 & 0.6145 & 0.6259 & 0.6633 & 0.6239 \\
        \midrule
        
        % Resnet18系列
        Resnet50-1shot        & 0.6884 & 0.7965 & 0.6831  & 0.3821 & 0.5187 & 0.3147  & 0.7380 & 0.8133 & 0.7178 & 0.8106 & 0.8306 & 0.8010 \\
        Resnet50-3shot        & 0.7070 & 0.7639 & 0.6831  & 0.8773 & 0.8921 & 0.8736  & 0.8741 & 0.8940 & 0.8626 & 0.7311 & 0.7852 & 0.6976 \\
        Resnet50-5shot        & 0.8602 & 0.8795 & 0.8575  & 0.7215 & 0.7977 & 0.7139  & 0.6676 & 0.7600 & 0.6171 & 0.7054 & 0.7632 & 0.6876 \\
        Resnet50-7shot        & 0.8709 & 0.8690 & 0.8568  & 0.9146 & 0.9260 & 0.9134  & 0.8591 & 0.8567 & 0.8362 & 0.9050 & 0.9146 & 0.9058 \\
        \midrule
        Syn-Diag-Cloud-1shot  & 0.9809 & 0.9815  & 0.9810  & 0.9722 & 0.9740  & 0.9723  & 0.9635 & 0.9664  & 0.9630 & 0.9395 & 0.9478 & 0.9391 \\
        Syn-Diag-Cloud-3shot  & 0.9869 & 0.9871  & 0.9869  & 0.9869 & 0.9877  & 0.9869  & 0.9771 & 0.9783  & 0.9770 & 0.9722 & 0.9745 & 0.9721 \\
        Syn-Diag-Cloud-5shot  & \textbf{0.9880} & \textbf{0.9883}  & \textbf{0.9880}  & 0.9924 & 0.9925  & 0.9924  & 0.9755 & 0.9760  & 0.9753 & 0.9733 & 0.9765 & 0.9729 \\
        Syn-Diag-Cloud-7shot  & 0.9847 & 0.9848  & 0.9847  & 0.9924 & 0.9925  & 0.9924  & \underline{0.9826} & \underline{0.9831}  & \underline{0.9825} & 0.9766 & 0.9777 & 0.9765 \\
        \midrule
        % Syn-Diag-Edge系列
        Syn-Diag-Edge-1shot  & 0.9716 & 0.9733  & 0.9716  & 0.9079 & 0.9384  & 0.9026  & 0.9564 & 0.9594  & 0.9557 & 0.9264 & 0.9354 & 0.9257 \\
        Syn-Diag-Edge-3shot  & 0.9858 & 0.9861  & 0.9858  & 0.9771 & 0.9800  & 0.9772  & 0.9727 & 0.9742  & 0.9727 & \underline{0.9776} & 0.9784 & \underline{0.9775} \\
        Syn-Diag-Edge-5shot  & \textbf{0.9880} & \textbf{0.9883}  & \textbf{0.9880}  & \underline{0.9940} & \underline{0.9941}  & \underline{0.9940}  & 0.9782 & 0.9788  & 0.9781 & \underline{0.9776} & \underline{0.9795} & 0.9774 \\
        Syn-Diag-Edge-7shot  & \underline{0.9875} & \underline{0.9876}  & \underline{0.9874}  & \textbf{0.9945} & \textbf{0.9946}  & \textbf{0.9945}  & \textbf{0.9847} & \textbf{0.9853}  & \textbf{0.9847} & \textbf{0.9815} & \textbf{0.9824} & \textbf{0.9814} \\
        \bottomrule
    \end{tabular}
}
\end{table}

%% file: table4.tex
\begin{table}[!t]
\caption{The training and test data on the SEU dataset come from the fault diagnosis analysis and evaluation of the 5-way task under the same working conditions (differentiated by row: 1-shot/3-shot/5-shot/7-shot)}
\label{tab:seu_results}
\centering

% 1. 颜色梯度定义（重点细分0.9-1.0区间，共7个梯度，保证渐变连贯）
% 低数值区（0.1-0.9）保留原逻辑，高数值区（0.9-1.0）拆分为3个细梯度
\definecolor{heat1}{rgb}{0.8, 0.9, 1.0}   % 浅蓝（0.1-0.3）→ 最差
\definecolor{heat2}{rgb}{0.8, 0.85, 0.95} % 浅青蓝（0.3-0.5）→ 较差
\definecolor{heat3}{rgb}{0.95, 0.85, 0.95}% 浅紫（0.5-0.7）→ 中等
\definecolor{heat4}{rgb}{1.0, 0.85, 0.85} % 浅粉红（0.7-0.9）→ 较好
\definecolor{heat5a}{rgb}{1.0, 0.8, 0.8}  % 浅红1（0.90-0.93）→ 最优-级
\definecolor{heat5b}{rgb}{1.0, 0.75, 0.75}% 浅红2（0.93-0.96）→ 最优-级
\definecolor{heat5c}{rgb}{1.0, 0.7, 0.7}  % 浅红3（0.96-1.00）→ 最优-级

\scalebox{0.8}{ % 保持表格宽度与可读性平衡
    \begin{tabular}{lcccccc}
        \toprule
        & \multicolumn{3}{c}{SEU-20} & \multicolumn{3}{c}{SEU-30} \\
        \cmidrule(lr){2-4} \cmidrule(lr){5-7}
        Methods    & ACC   & Precision      & F1     & ACC  & Precision   & F1 \\
        \midrule
        % -------------------------- CNN系列 --------------------------
        CNN-1shot         & \cellcolor{heat4}0.7646 & \cellcolor{heat4}0.8022 & \cellcolor{heat4}0.7612  & \cellcolor{heat2}0.4813 & \cellcolor{heat2}0.4257 & \cellcolor{heat2}0.4083  \\
        CNN-3shot         & \cellcolor{heat2}0.4188 & \cellcolor{heat1}0.2688 & \cellcolor{heat1}0.3177  & \cellcolor{heat2}0.4542 & \cellcolor{heat2}0.3379 & \cellcolor{heat2}0.3544  \\
        CNN-5shot         & \cellcolor{heat3}0.6521 & \cellcolor{heat3}0.6269 & \cellcolor{heat3}0.6186  & \cellcolor{heat2}0.3396 & \cellcolor{heat1}0.1380 & \cellcolor{heat1}0.1951  \\
        CNN-7shot         & \cellcolor{heat4}0.8063 & \cellcolor{heat4}0.8375 & \cellcolor{heat4}0.7937  & \cellcolor{heat2}0.3479 & \cellcolor{heat1}0.1404 & \cellcolor{heat1}0.1998  \\
        \midrule
        
        % -------------------------- VisionTransformer系列 --------------------------
        ViT-1shot & \cellcolor{heat3}0.6458 & \cellcolor{heat3}0.5393 & \cellcolor{heat3}0.5850  & \cellcolor{heat3}0.5875 & \cellcolor{heat3}0.5678 & \cellcolor{heat3}0.5622  \\
        ViT-3shot & \cellcolor{heat2}0.4000 & \cellcolor{heat1}0.2025 & \cellcolor{heat1}0.2560  & \cellcolor{heat2}0.5000 & \cellcolor{heat2}0.3162 & \cellcolor{heat2}0.3823  \\
        ViT-5shot & \cellcolor{heat1}0.2417 & \cellcolor{heat1}0.1094 & \cellcolor{heat1}0.1431  & \cellcolor{heat2}0.4938 & \cellcolor{heat2}0.3414 & \cellcolor{heat2}0.3746  \\
        ViT-7shot & \cellcolor{heat3}0.5479 & \cellcolor{heat3}0.5436 & \cellcolor{heat2}0.4913  & \cellcolor{heat3}0.5583 & \cellcolor{heat3}0.5490 & \cellcolor{heat3}0.5476  \\
        \midrule
        % -------------------------- Resnet50系列 --------------------------
        Resnet50-1shot    & \cellcolor{heat4}0.7833 & \cellcolor{heat4}0.7720 & \cellcolor{heat4}0.7716  & \cellcolor{heat3}0.5979 & \cellcolor{heat3}0.5717 & \cellcolor{heat3}0.5434  \\
        Resnet50-3shot    & \cellcolor{heat3}0.6708 & \cellcolor{heat3}0.5745 & \cellcolor{heat3}0.6030  & \cellcolor{heat3}0.5667 & \cellcolor{heat3}0.5918 & \cellcolor{heat3}0.5032  \\
        Resnet50-5shot    & \cellcolor{heat4}0.7688 & \cellcolor{heat4}0.7746 & \cellcolor{heat4}0.7680  & \cellcolor{heat3}0.6604 & \cellcolor{heat3}0.7003 & \cellcolor{heat3}0.6131  \\
        Resnet50-7shot    & \cellcolor{heat4}0.7896 & \cellcolor{heat4}0.8642 & \cellcolor{heat4}0.7605  & \cellcolor{heat4}0.8458 & \cellcolor{heat4}0.8538 & \cellcolor{heat4}0.8402  \\
        \midrule
        % -------------------------- Syn-Diag-Cloud系列（细分0.9-1.0颜色） --------------------------
        Syn-Diag-Cloud-1shot  & \cellcolor{heat5a}0.9167 & \cellcolor{heat5a}0.9188 & \cellcolor{heat5a}0.9158  & \cellcolor{heat5b}0.9479 & \cellcolor{heat5b}0.9536 & \cellcolor{heat5b}0.9487  \\
        Syn-Diag-Cloud-3shot  & \cellcolor{heat5a}0.9250 & \cellcolor{heat5a}0.9257 & \cellcolor{heat5a}0.9248  & \cellcolor{heat5c}0.9646 & \cellcolor{heat5c}0.9649 & \cellcolor{heat5c}0.9644  \\
        Syn-Diag-Cloud-5shot  & \cellcolor{heat5a}0.9292 & \cellcolor{heat5a}0.9306 & \cellcolor{heat5a}0.9286  & \cellcolor{heat5c}0.9688 & \cellcolor{heat5c}0.9694 & \cellcolor{heat5c}0.9688  \\
        Syn-Diag-Cloud-7shot  & \cellcolor{heat5b}0.9417 & \cellcolor{heat5b}0.9424 & \cellcolor{heat5b}0.9411  & \cellcolor{heat5c}0.9688 & \cellcolor{heat5c}0.9691 & \cellcolor{heat5c}0.9687  \\
        \midrule
        % -------------------------- Syn-Diag-Edge系列（细分0.9-1.0颜色） --------------------------
        Syn-Diag-Edge-1shot  & \cellcolor{heat5a}0.9271 & \cellcolor{heat5a}0.9292 & \cellcolor{heat5a}0.9261  & \cellcolor{heat5b}0.9458 & \cellcolor{heat5b}0.9479 & \cellcolor{heat5b}0.9462  \\
        Syn-Diag-Edge-3shot  & \cellcolor{heat4}0.8896  & \cellcolor{heat4}0.8985  & \cellcolor{heat4}0.8911  & \cellcolor{heat5c}0.9604 & \cellcolor{heat5c}0.9610 & \cellcolor{heat5c}0.9603  \\
        Syn-Diag-Edge-5shot  & \cellcolor{heat5b}0.9354 & \cellcolor{heat5b}0.9365 & \cellcolor{heat5b}0.9347  & \cellcolor{heat5c}0.9708 & \cellcolor{heat5c}0.9713 & \cellcolor{heat5c}0.9709  \\
        Syn-Diag-Edge-7shot  & \cellcolor{heat5b}0.9396 & \cellcolor{heat5b}0.9408 & \cellcolor{heat5b}0.9387  & \cellcolor{heat5c}0.9729 & \cellcolor{heat5c}0.9731 & \cellcolor{heat5c}0.9729  \\
        \bottomrule
    \end{tabular}
}

% \vspace{8pt}
% % 不再使用 scalebox，而是组合使用 small 和 setlength
% \centering
% \small % 1. 先将基础字号缩小一级
% { % 2. 使用大括号创建一个局部作用域
%   \setlength{\tabcolsep}{3pt} % 3. 在局部作用域内将列间距减小（可尝试 2pt, 3pt, 4pt）
%   \begin{tabular}{ccccccc} % 注意：您的原始代码是cccccccc（8个c），这里修正为7个
%     % 颜色块
%     \cellcolor{heat1}\rule{12pt}{8pt} & 
%     \cellcolor{heat2}\rule{12pt}{8pt} & 
%     \cellcolor{heat3}\rule{12pt}{8pt} & 
%     \cellcolor{heat4}\rule{12pt}{8pt} & 
%     \cellcolor{heat5a}\rule{12pt}{8pt} & 
%     \cellcolor{heat5b}\rule{12pt}{8pt} & 
%     \cellcolor{heat5c}\rule{12pt}{8pt} \\
%     % 对应数值区间
%     0.1-0.3 & 0.3-0.5 & 0.5-0.7 & 0.7-0.9 & 0.90-0.93 & 0.93-0.96 & 0.96-1.00 \\
%     % 解读说明 (修改之处)
%     \multicolumn{7}{>{\raggedright\arraybackslash}p{\columnwidth}}{\small Note: The color transitions from light blue to light red. The darker the color, the better the performance within the same optimal echelon.}
%   \end{tabular}
% }
% --- 定义一个新的、内容水平居中的 X 列类型 ---
\newcolumntype{C}{>{\centering\arraybackslash}X}
\vspace{8pt}
\centering
{
  % ======================== 核心修改在这里 ========================
  % 1. 将列间距设置为 0
  \setlength{\tabcolsep}{0pt}
  
  % 2. 使用标准 tabular，并用 @{} 移除列之间的所有空白
  \begin{tabular}{*{7}{c@{\hspace{0pt}}}} % 使用 c 列，并用 @{} 确保无缝连接
    % 第一行：无缝连接的色块
    {\color{heat1}\rule{40pt}{8pt}} & 
    {\color{heat2}\rule{40pt}{8pt}} & 
    {\color{heat3}\rule{40pt}{8pt}} & 
    {\color{heat4}\rule{40pt}{8pt}} & 
    {\color{heat5a}\rule{40pt}{8pt}} & 
    {\color{heat5b}\rule{40pt}{8pt}} & 
    {\color{heat5c}\rule{40pt}{8pt}} \\
    
    % 第二行：数值区间 (现在会在各自20pt的宽度内居中)
    \scriptsize 0.10-0.30 & \scriptsize 0.30-0.50 & \scriptsize 0.50-0.70 & \scriptsize 0.70-0.90 & \scriptsize 0.90-0.93 & \scriptsize 0.93-0.96 & \scriptsize 0.96-1.00 \\
  \end{tabular}
  % =============================================================
  
  % 第三行说明可以单独处理，以获得更好的布局
  \par % 另起一段
  \vspace{2pt} % 稍微增加一点垂直间距
  \begin{minipage}{\linewidth} % 创建一个与文本等宽的小环境
    \raggedright % 左对齐
    \small Note: The color transitions from light blue to light red. The darker the color, the better the performance within the same optimal echelon.
  \end{minipage}

}
\end{table}

%% file: table3.tex
\begin{table}[!t]
\centering
\caption{Cross-condition generalization performance evaluation on the CWRU dataset. Experiments were conducted in a 19-way 5-shot setting, where the notation "A -> B" indicates that the model is fine-tuned on the training set of source condition A (e.g., 1750R) and then evaluated on the test set of target condition B (e.g., 1730R) to verify the model's generalization ability.}
\label{tab:cross_domain_results}
\scalebox{0.8}{ % Adjust the scale to fit the page
    \begin{tabular}{lcccccccccccc}
        \toprule
        & \multicolumn{3}{c}{1730R -> 1750R} & \multicolumn{3}{c}{1750R -> 1730R} & \multicolumn{3}{c}{1772R -> 1797R} & \multicolumn{3}{c}{1797R -> 1772R} \\
        \cmidrule(lr){2-4} \cmidrule(lr){5-7} \cmidrule(lr){8-10} \cmidrule(lr){11-13}
        Methods   & ACC   & Precision & F1     & ACC   & Precision & F1       & ACC   & Precision & F1     & ACC   & Precision & F1 \\
        \midrule
        Syn-Diag-Cloud  & 0.9651 & 0.9702 & 0.9645 &  0.9738 & 0.9755 & 0.9739 &0.9597 & 0.9639 & 0.9598 & 0.9711 & 0.9730 &  0.9711 \\
        \midrule
        & \multicolumn{3}{c}{1730R -> 1772R} & \multicolumn{3}{c}{1750R -> 1797R} & \multicolumn{3}{c}{1772R -> 1730R} & \multicolumn{3}{c}{1797R -> 1750R} \\
        \cmidrule(lr){2-4} \cmidrule(lr){5-7} \cmidrule(lr){8-10} \cmidrule(lr){11-13}
        Methods   & ACC   & Precision & F1     & ACC   & Precision & F1       & ACC   & Precision & F1     & ACC   & Precision & F1 \\
        \midrule
        Syn-Diag-Cloud  & 0.9684 & 0.9708 & 0.9685 &0.9733 & 0.9757 &  0.9735 &  0.9733 &  0.9755 & 0.9733 & 0.9771 & 0.9792 & 0.9771 \\
        \midrule
        & \multicolumn{3}{c}{1730R -> 1797R} & \multicolumn{3}{c}{1750R -> 1772R} & \multicolumn{3}{c}{1772R -> 1750R} & \multicolumn{3}{c}{1797R -> 1730R} \\
        \cmidrule(lr){2-4} \cmidrule(lr){5-7} \cmidrule(lr){8-10} \cmidrule(lr){11-13}
        Methods   & ACC   & Precision & F1     & ACC   & Precision & F1       & ACC   & Precision & F1     & ACC   & Precision & F1 \\
        \midrule
        Syn-Diag-Cloud  & 0.9771 &  0.9788 & 0.9772 &  0.9804 & 0.9812 & 0.9804 & 0.9858 & 0.9861 & 0.9858 & 0.9673 &0.9699 & 0.9674 \\
        \bottomrule
    \end{tabular}
}
\end{table}

%% file: Conclusion.tex
\section{Conclusion}

This paper introduced Syn-Diag, an end-to-end framework bridging the gap between Large Language Models and the practical constraints of industrial fault diagnosis. Our primary contribution is a synergistic architecture addressing key industry challenges. Through visual-semantic alignment, we significantly reduce data dependency, achieving remarkable few-shot and cross-condition performance. A dynamic reasoning process boosts discriminative power, while our cloud-edge architecture with knowledge distillation enables efficient deployment on resource-constrained devices without performance loss. The framework's sustainability is ensured by a low-cost online updating mechanism. In essence, Syn-Diag offers a validated paradigm for deploying powerful, adaptive, and efficient diagnostic systems. Future work will extend this framework to more sensor modalities and explore advanced collaborative strategies.

%% file: templateArxiv.bbl
\begin{thebibliography}{10}

\bibitem{zhu2023review}
Zhiqin Zhu, Yangbo Lei, Guanqiu Qi, Yi~Chai, Neal Mazur, Yiyao An, and Xinghua Huang.
\newblock A review of the application of deep learning in intelligent fault diagnosis of rotating machinery.
\newblock {\em Measurement}, 206:112346, 2023.

\bibitem{zhou2022towards}
Taotao Zhou, Te~Han, and Enrique~Lopez Droguett.
\newblock Towards trustworthy machine fault diagnosis: A probabilistic bayesian deep learning framework.
\newblock {\em Reliability Engineering \& System Safety}, 224:108525, 2022.

\bibitem{qin2022fault}
Yufeng Qin and Xianjun Shi.
\newblock Fault diagnosis method for rolling bearings based on two-channel cnn under unbalanced datasets.
\newblock {\em Applied Sciences}, 12(17):8474, 2022.

\bibitem{yin2024bi}
Linfei Yin and Zixuan Wang.
\newblock Bi-level binary coded fully connected classifier based on residual network 50 with bottom and deep level features for bearing fault diagnosis.
\newblock {\em Engineering Applications of Artificial Intelligence}, 133:108342, 2024.

\bibitem{ji2024devit}
Mintong Ji and Guodong Zhao.
\newblock Devit: Deformable convolution based-vision transformer for bearing fault diagnosis.
\newblock {\em IEEE Transactions on Instrumentation and Measurement}, 2024.

\bibitem{jiang2025adaptive}
Yonghua Jiang, Maoli Lu, Zhilin Dong, Zhichao Jiang, Weidong Jiao, Chao Tang, Jianfeng Sun, and Zhongyi Xuan.
\newblock Adaptive deeping siamese residual network: A novel model for few-shot bearing fault diagnosis.
\newblock {\em Machines}, 13(3):193, 2025.

\bibitem{pan2024cloud}
Yanghe Pan, Zhou Su, Yuntao Wang, Shaolong Guo, Han Liu, Ruidong Li, and Yuan Wu.
\newblock Cloud-edge collaborative large model services: Challenges and solutions.
\newblock {\em IEEE Network}, 2024.

\bibitem{he2025full}
David He, Miao He, and Jay Yoon.
\newblock Full ceramic bearing fault diagnosis with few-shot learning using gpt-2.
\newblock {\em Computer Modeling in Engineering \& Sciences (CMES)}, 143(2), 2025.

\bibitem{zhao2025adjust}
Jiancheng Zhao, Jiaqi Yue, Chunhui Zhao, and Chen Chen.
\newblock Adjust to reality: Llm-driven test-time semantic adjustment for zero-shot fault diagnosis.
\newblock {\em Control Engineering Practice}, 164:106406, 2025.

\bibitem{ran2025fd}
Guangsheng Ran, Xueyi Li, Ming Li, Qi~Li, Tianyang Wang, and Fulei Chu.
\newblock Fd-zero framework: Llm-based diagnosis framework for zero-shot mechanical time-domain signals.
\newblock {\em Structural Health Monitoring}, page 14759217251366940, 2025.

\bibitem{lin2025fd}
Lin Lin, Sihao Zhang, Song Fu, and Yikun Liu.
\newblock Fd-llm: Large language model for fault diagnosis of complex equipment.
\newblock {\em Advanced Engineering Informatics}, 65:103208, 2025.

\bibitem{zheng2024empirical}
Shuwen Zheng, Kai Pan, Jie Liu, and Yunxia Chen.
\newblock Empirical study on fine-tuning pre-trained large language models for fault diagnosis of complex systems.
\newblock {\em Reliability Engineering \& System Safety}, 252:110382, 2024.

\bibitem{xiao2025krail}
Xingyu Xiao, Peng Chen, Ben Qi, Hongru Zhao, Jingang Liang, Jiejuan Tong, and Haitao Wang.
\newblock Krail: A knowledge-driven framework for human reliability analysis integrating idheas-data and large language models.
\newblock {\em Reliability Engineering \& System Safety}, page 111585, 2025.

\bibitem{wang2025multi}
Tian Wang, Ping Wang, Feng Yang, Shuai Wang, Qiang Fang, and Meng Chi.
\newblock Multi large language model collaboration framework for few-shot link prediction in evolutionary fault diagnosis event graphs.
\newblock {\em Journal of Process Control}, 145:103342, 2025.

\bibitem{radford2021learning}
Alec Radford, Jong~Wook Kim, Chris Hallacy, Aditya Ramesh, Gabriel Goh, Sandhini Agarwal, Girish Sastry, Amanda Askell, Pamela Mishkin, Jack Clark, et~al.
\newblock Learning transferable visual models from natural language supervision.
\newblock In {\em International conference on machine learning}, pages 8748--8763. PMLR, 2021.

\bibitem{wang2025diagllm}
Jie Wang, Tianrui Li, Yan Yang, Shiqian Chen, and Wanming Zhai.
\newblock Diagllm: multimodal reasoning with large language model for explainable bearing fault diagnosis.
\newblock {\em Science China Information Sciences}, 68(6):160103, 2025.

\bibitem{zhao2020deep}
Zhibin Zhao, Tianfu Li, Jingyao Wu, Chuang Sun, Shibin Wang, Ruqiang Yan, and Xuefeng Chen.
\newblock Deep learning algorithms for rotating machinery intelligent diagnosis: An open source benchmark study.
\newblock {\em ISA transactions}, 107:224--255, 2020.

\bibitem{li2022multi}
Wei Li, Xiang Zhong, Haidong Shao, Baoping Cai, and Xingkai Yang.
\newblock Multi-mode data augmentation and fault diagnosis of rotating machinery using modified acgan designed with new framework.
\newblock {\em Advanced Engineering Informatics}, 52:101552, 2022.

\bibitem{sifat2025gan}
Md~Sulyman~Islam Sifat, Md~Alamgir Kabir, MM~Manjurul Islam, Atiq~Ur Rehman, and Amine Bermak.
\newblock Gan-based data augmentation for fault diagnosis and prognosis of rolling bearings: A literature review.
\newblock {\em IEEE Access}, 2025.

\bibitem{an2023gaussian}
Yiyao An, Ke~Zhang, Yi~Chai, Zhiqin Zhu, and Qie Liu.
\newblock Gaussian mixture variational-based transformer domain adaptation fault diagnosis method and its application in bearing fault diagnosis.
\newblock {\em IEEE Transactions on Industrial Informatics}, 20(1):615--625, 2023.

\bibitem{xia2025fault}
Huaitao Xia, Tao Meng, Zonglin Zuo, and Wenjie Ma.
\newblock Fault semantic knowledge transfer learning: Cross-domain compound fault diagnosis method under limited single fault samples.
\newblock {\em Reliability Engineering \& System Safety}, 260:111050, 2025.

\bibitem{zhang2025adaptive}
Xiao Zhang, Weiguo Huang, Jun Wang, Zhongkui Zhu, Changqing Shen, Kai Chen, Xingli Zhong, and Li~He.
\newblock Adaptive variational sampling-embedded domain generalization network for fault diagnosis with intra-inter-domain class imbalance.
\newblock {\em Reliability Engineering \& System Safety}, 256:110707, 2025.

\bibitem{zhu2020new}
Jun Zhu, Nan Chen, and Changqing Shen.
\newblock A new multiple source domain adaptation fault diagnosis method between different rotating machines.
\newblock {\em IEEE Transactions on Industrial Informatics}, 17(7):4788--4797, 2020.

\bibitem{qian2024adaptive}
Quan Qian, Jun Luo, and Yi~Qin.
\newblock Adaptive intermediate class-wise distribution alignment: A universal domain adaptation and generalization method for machine fault diagnosis.
\newblock {\em IEEE transactions on neural networks and learning systems}, 36(3):4296--4310, 2024.

\bibitem{che2022few}
Changchang Che, Huawei Wang, Minglan Xiong, and Xiaomei Ni.
\newblock Few-shot fault diagnosis of rolling bearing under variable working conditions based on ensemble meta-learning.
\newblock {\em Digital Signal Processing}, 131:103777, 2022.

\bibitem{liang2025interpretable}
Shuang Liang, Jinsong Yu, Diyin Tang, and Xu~Ke.
\newblock Interpretable attention-based prototype network for uav fault diagnosis under small sample conditions.
\newblock {\em Reliability Engineering \& System Safety}, page 111601, 2025.

\bibitem{zhang2022supervised}
Yongchao Zhang, Zhaohui Ren, Shihua Zhou, Ke~Feng, Kun Yu, and Zheng Liu.
\newblock Supervised contrastive learning-based domain adaptation network for intelligent unsupervised fault diagnosis of rolling bearing.
\newblock {\em IEEE/ASME Transactions on Mechatronics}, 27(6):5371--5380, 2022.

\bibitem{wu2020few}
Jingyao Wu, Zhibin Zhao, Chuang Sun, Ruqiang Yan, and Xuefeng Chen.
\newblock Few-shot transfer learning for intelligent fault diagnosis of machine.
\newblock {\em Measurement}, 166:108202, 2020.

\bibitem{brown2020language}
Tom Brown, Benjamin Mann, Nick Ryder, Melanie Subbiah, Jared~D Kaplan, Prafulla Dhariwal, Arvind Neelakantan, Pranav Shyam, Girish Sastry, Amanda Askell, et~al.
\newblock Language models are few-shot learners.
\newblock {\em Advances in neural information processing systems}, 33:1877--1901, 2020.

\bibitem{liu2023visual}
Haotian Liu, Chunyuan Li, Qingyang Wu, and Yong~Jae Lee.
\newblock Visual instruction tuning.
\newblock {\em Advances in neural information processing systems}, 36:34892--34916, 2023.

\bibitem{dai2023instructblip}
Wenliang Dai, Junnan Li, Dongxu Li, Anthony Tiong, Junqi Zhao, Weisheng Wang, Boyang Li, Pascale~N Fung, and Steven Hoi.
\newblock Instructblip: Towards general-purpose vision-language models with instruction tuning.
\newblock {\em Advances in neural information processing systems}, 36:49250--49267, 2023.

\bibitem{hu2022lora}
Edward~J Hu, Yelong Shen, Phillip Wallis, Zeyuan Allen-Zhu, Yuanzhi Li, Shean Wang, Lu~Wang, Weizhu Chen, et~al.
\newblock Lora: Low-rank adaptation of large language models.
\newblock {\em ICLR}, 1(2):3, 2022.

\bibitem{peifeng2024joint}
LIU Peifeng, Lu~Qian, Xingwei Zhao, and Bo~Tao.
\newblock Joint knowledge graph and large language model for fault diagnosis and its application in aviation assembly.
\newblock {\em IEEE Transactions on Industrial Informatics}, 20(6):8160--8169, 2024.

\bibitem{yang2024large}
Aidan~ZH Yang, Claire Le~Goues, Ruben Martins, and Vincent Hellendoorn.
\newblock Large language models for test-free fault localization.
\newblock In {\em Proceedings of the 46th IEEE/ACM International Conference on Software Engineering}, pages 1--12, 2024.

\bibitem{tao2025llm}
Laifa Tao, Haifei Liu, Guoao Ning, Wenyan Cao, Bohao Huang, and Chen Lu.
\newblock Llm-based framework for bearing fault diagnosis.
\newblock {\em Mechanical Systems and Signal Processing}, 224:112127, 2025.

\bibitem{smith2015rolling}
Wade~A Smith and Robert~B Randall.
\newblock Rolling element bearing diagnostics using the case western reserve university data: A benchmark study.
\newblock {\em Mechanical systems and signal processing}, 64:100--131, 2015.

\bibitem{shao2018highly}
Siyu Shao, Stephen McAleer, Ruqiang Yan, and Pierre Baldi.
\newblock Highly accurate machine fault diagnosis using deep transfer learning.
\newblock {\em IEEE transactions on industrial informatics}, 15(4):2446--2455, 2018.

\end{thebibliography}
